\documentclass[runningheads]{llncs}
\usepackage{booktabs}
\usepackage[T1]{fontenc}
\usepackage[section]{placeins}
\usepackage{float}
\usepackage{multirow}
\usepackage{makecell} 
\usepackage{url}
\usepackage{hyperref}
\usepackage{gensymb}
\usepackage{graphicx}
\usepackage{subcaption}
\usepackage{graphicx}     
\usepackage{subcaption}   
\usepackage{float}  
\usepackage{graphicx}
\usepackage{tikz}
\usepackage{import}
\usetikzlibrary{positioning,3d,calc}
\usepackage{multirow}
\usepackage{graphicx}
\usepackage{tabularx}
\usepackage{booktabs}
\usepackage{import}
\usepackage{svg}
\usepackage{amsmath}
\usepackage{amssymb}
\usepackage{pgfplots}
\pgfplotsset{compat=1.17}
\usepackage{pgfplotstable}
\usepackage{xcolor}
\usepackage[flushleft]{threeparttable}

\begin{document}
{
\title{Learning Regional Monsoon Patterns with a Multimodal Attention U-Net}

\author{
Swaib Ilias Mazumder\inst{1} \and
Manish Kumar\inst{2} \and
Aparajita Khan\inst{3}
}

\institute{
\textsuperscript{1}Computer Science \& Engineering, Indian Institute of Technology Roorkee, India\\
\email{swaib\_m@cs.iitr.ac.in}\\
\textsuperscript{2}Computer Science \& Engineering, Indian Institute of Technology Ropar, India\\
\email{manishk@iitrpr.ac.in} \\ 
\textsuperscript{3} Computer Science \& Engineering, Indian Institute of Technology (BHU) Varanasi, India\\
\email{aparajita.cse@iitbhu.ac.in}
} 
\maketitle            

\begin{abstract}
Accurate long-range monsoon rainfall prediction is critical for India’s rain-fed agricultural economy and climate resilience planning, yet remains hindered by sparse ground data and complex regional variability. This work proposes a multimodal deep learning framework for gridded precipitation classification using satellite-derived geospatial inputs. Unlike previous rainfall prediction methods relying on coarse-resolution datasets of 5–50 km grid, we curate a high-resolution dataset of projected 1 km grid resolution for five Indian states, integrating seven heterogeneous Earth observation modalities, including land surface temperature, vegetation, soil moisture, humidity, wind speed, elevation, and land use, spanning the June–September 2024 period. We adopt a attention-guided U-Net architecture that captures spatial patterns and temporal dependencies across multi-modalities, and propose a combination of focal and dice loss to address class imbalance and spatial coherence in rainfall categories defined by the India Meteorological Department. Extensive experiments show that the multi-model framework significantly outperforms unimodal baselines and existing deep approaches, especially in underrepresented extreme rainfall zones. The framework demonstrates potential for scalable, region-adaptive monsoon forecasting and Earth observation driven climate risk assessment.
\end{abstract}

\keywords{Multimodal Earth Observation \and Precipitation Forecasting \and Spatio-Temporal Modeling \and Remote Sensing Data Fusion \and Attention U-Net}}

\section{Introduction}

India's climate system exhibits pronounced spatial variability, with regions experiencing diverse precipitation patterns and climate-related risks, ranging from droughts in the central plains to floods in the northeast and erratic rainfall along the coasts \cite{Dash2024SpatialPrecip,Roxy2017CentralIndia,Ghosh2011RainfallExtremes}. These extremes disrupt agriculture, water management, and disaster preparedness. Precipitation is driven by a complex interplay of climatic and environmental factors, including temperature, humidity, wind, solar radiation, and aerosols \cite{Hou2014GPM}. Despite the availability of these rich Earth observation data from satellites (e.g., MODIS, Sentinel, INSAT) and in-situ rain gauge networks, India continues to lack scalable, high-resolution approaches for region-specific precipitation prediction \cite{Maity2013HydrologyMonsoon,Tirumani2021Intercomp}.

Short-term forecasts (nowcasting) with lead times of minutes to hours are effective for real-time warnings in flood control and aviation \cite{shi2017deep,lam2022learning}. However, in a predominantly rain-fed agro-economy like India, long-range forecasting with lead times of 30 to 90 days is crucial for seasonal risk management and agricultural planning \cite{skilful}. These forecasts often rely on Long Period Average (LPA) based assessments, which quantify rainfall deviations from climatological norms \cite{imd2024monsoon}. Rainfall is typically categorized into classes such as \textit{Scarcity}, \textit{Deficit}, \textit{Normal}, and \textit{Excess}, which inform official monsoon outlooks and guide decisions related to crop planning and water management.

Current forecasts by the Indian Meteorological Department (IMD) primarily rely on numerical weather prediction models incorporating large-scale phenomena such as El Niño \cite{ebiak1987model}, the Indian Ocean Dipole \cite{inbook}, and the Monsoon Mission Coupled Forecasting System \cite{gmd-17-709-2024}. While effective at national and zonal scales, these models often struggle to capture fine-scale spatial heterogeneity and are increasingly challenged by evolving climate regimes. Recently, remote sensing driven machine learning and deep learning approaches have shown promise for rainfall classification, leveraging methods such as recurrent neural networks, graph neural networks, transformers, and ensemble techniques \cite{yu2023multisource,shi2017deep}. However, their deployment in the Indian context, particularly for high-resolution precipitation classification, remains limited. A central challenge lies in fusing heterogeneous multi-source spatio-temporal data such as LST, NDVI, soil moisture, wind speed, and relative humidity which differ in units, resolutions, and dynamics \cite{yu2023multisource}. For instance, LST may drop sharply from June to August due to cloud cover, while elevation remains static yet spatially diverse. Additionally, imbalanced precipitation classes, especially in hilly or coastal areas, further complicate modeling. Addressing these issues necessitates models that can capture nonlinear intra- and inter-modality dependencies while maintaining robustness to spatial and class distribution disparities.

This paper proposes a deep learning framework for gridded classification of precipitation categories using IMD’s LPA-based thresholds. In this work, ‘high-resolution’ refers to gridded satellite and reanalysis data harmonized to a spatial resolution of 1 km × 1 km, which is significantly finer compared to typical climate datasets (5 km – 50 km) used in previous rainfall forecasting studies. This higher spatial granularity allows more localized and region-specific precipitation prediction. A high-resolution dataset is curated for five ecologically diverse Indian states for the 2024 monsoon season, incorporating  a comprehensive set of seven heterogeneous predictor modalities: NDVI, LST, soil moisture, wind speed, humidity, elevation, and land-use/land-cover. The proposed model addresses the outlined challenges by leveraging multi-dimensional convolutions for spatial locality, attention mechanisms for modality fusion and spatio-temporal relevance, and spatially weighted loss functions to handle class imbalance while maintaining coherence in predicted maps. The effectiveness of the proposed model is evaluated against both unimodal and existing machine/deep learning baselines.

\section{Related Works}
Rainfall prediction has been extensively studied using machine learning, deep learning, and hybrid methods. Classical machine learning models, including random forests, support vector regression, and artificial neural networks, have been applied to regions like the Himalayas, northeastern and central India  \cite{Dash2024SpatialPrecip,NEIndiaExtremes2023,Roxy2017CentralIndia}, and semi-arid zones \cite{10.2166/hydro.2024.014}, shows improved performance over traditional statistical techniques. More recent works leverage deep learning architectures such as recurrent and convolution networks \cite{shi2017deep,ADERYANI2022128463} to better model sequential and spatial dependencies in rainfall events \cite{Wani2024,yu2023multisource,shi2017deep}.

Precipitation nowcasting has also advanced through the use of radar data combined with coarse-resolution climate variables from reanalysis products like ERA5. Methods incorporating ERA5 variables (e.g., precipitable water vapor, wind divergence, and atmospheric pressure) with neural networks \cite{XU20251732} or transformer based models \cite{nowcasting_cao} demonstrate the relevance of atmospheric features in rainfall forecasting. However, such inputs typically have coarse spatial resolutions (0.25°–0.5°, i.e., 25–50 km grids), limiting localized forecasting potential.
To address these limitations, multimodal frameworks integrating diverse satellite derived modalities have been explored in limited contexts. While radar based models exist \cite{rs17071123,Xu2025},  attention-guided multimodal neural networks operating on harmonized high-resolution spatio-temporal grids remain underexplored. Our work addresses this gap by developing a multimodal attention U-Net that fuses multiple geospatial inputs for localized precipitation classification.

\section{Methodology}
\begin{figure*}[t]
\vspace{-0.3cm}
    \centering
    \includegraphics[trim=0.5cm 0.4cm 0.8cm 1cm, clip, scale = 0.46]{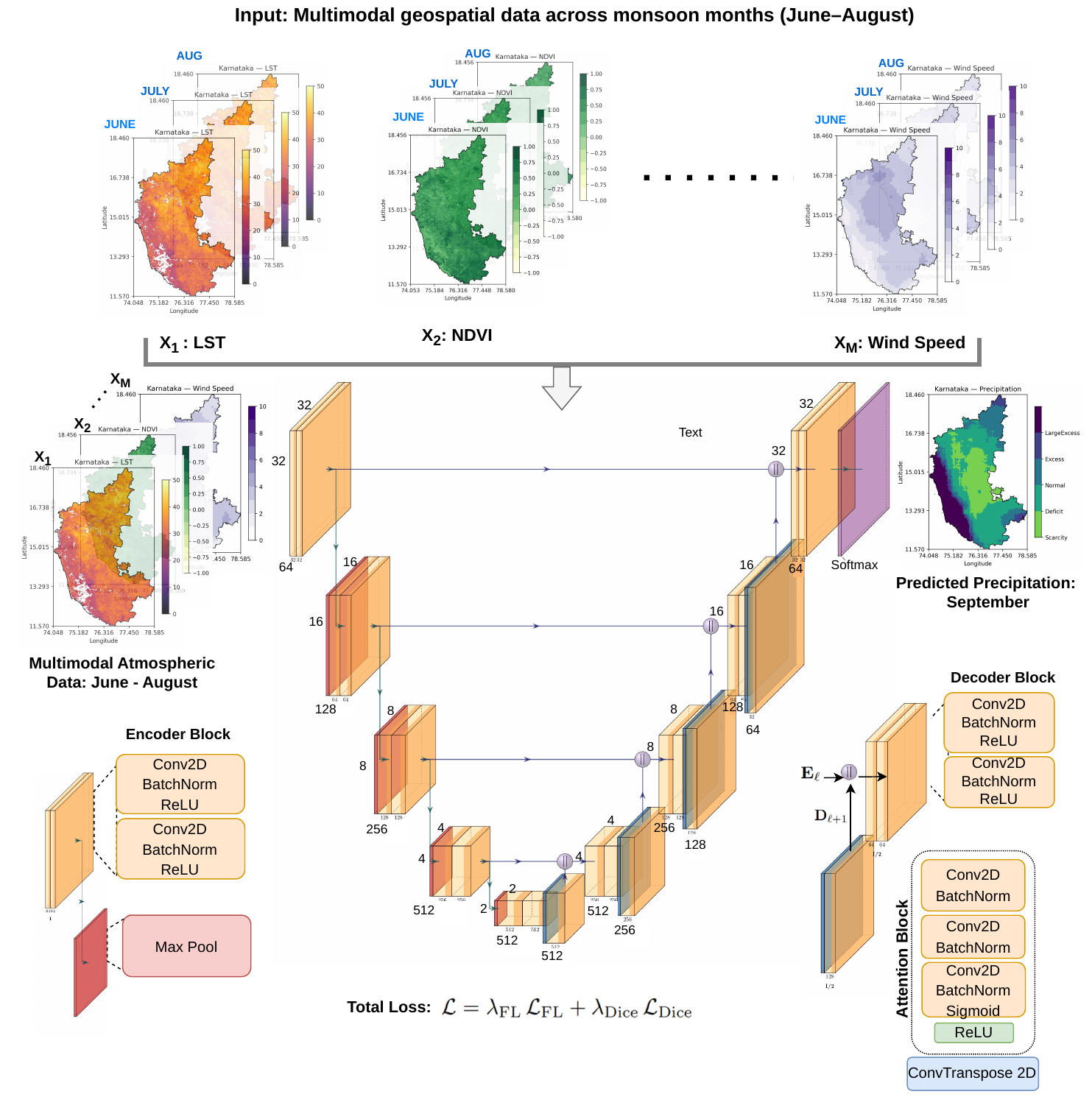}
    \caption{\small Proposed Multimodal Attention U-Net for Grid-wise Precipitation Classification.
    Multiple geospatial inputs (e.g., LST, NDVI, Wind Speed) over June–August are encoded via an attention U-Net to produce spatially discriminative features that can forecast categorical precipitation for September.}
    \label{fig:mvforecast}
\vspace{-0.5cm}
\end{figure*}

Given a set of gridded spatio-temporal observations from $M$ satellite-derived modalities (e.g., vegetation indices, soil moisture, LST) collected over a region at $T$ time steps on an $H \times W$ grid, the objective is to predict the categorical precipitation map (e.g., `Deficit', `Normal', and `Excess')for a future time point $\Delta$. The input is defined as
\begin{equation}
\mathcal{X} = \{ X_1, X_2, \ldots, X_M \}, \quad X_m \in \mathbb{R}^{H \times W \times T},
\label{eq:mvData}
\end{equation}
where each $X_m$ is a spatio-temporal tensor corresponding to the $m$-th modality, and $\mathcal{X}$ provides a rich multimodal representation of atmospheric dynamics. The target output is the precipitation category map at time $T+\Delta$, given by 
$Y^{T+\Delta} \in \{1, 2, \ldots, K\}^{H \times W},$
where $K$ is the number of discrete precipitation classes, and $Y^{T+\Delta}(i,j)$ denotes the predicted class at grid cell $(i,j)$.

The proposed approach extends the Attention U-Net architecture~\cite{atten_unet}, originally developed for medical image segmentation, to the domain of multimodal spatio-temporal rainfall classification. Building on the spatial attention mechanism at its core, we introduce key modifications to address the challenges of modeling precipitation over climatically diverse regions. Specifically, our contributions include: (i) a fusion strategy that integrates temporally stacked, multimodal geospatial inputs for effective representation learning; (ii) a customized composite loss function to address regional class imbalance and enhance spatial coherence; and (iii) application of the model to a curated, high-resolution dataset covering five Indian states.

\subsection{Multimodal Spatio-Temporal Data Fusion}
A robust precipitation forecasting framework must capture spatial, temporal, and cross-modal dependencies from heterogeneous geospatial inputs. To this end, we propose a multimodal attention-based neural network that jointly encodes spatio-temporal features via a shared encoder, with attention-guided skip connections preserving precipitation-relevant signals. 

Given the modality-specific input tensors $\mathcal{X}$ from Eq. (~\ref{eq:mvData}), a unified composite tensor $\mathbf{X} \in \mathbb{R}^{H \times W \times C}$ is formed by stacking temporal slices from all $M$ modalities along the channel dimension, where $C = M \times T$. This is represented as:
\begin{equation}
  \mathbf{X} = \left[\, \mathbf{X}_1^{(1{:}T)} \;\middle|\; \mathbf{X}_2^{(1{:}T)} \;\middle|\; \cdots \;\middle|\; \mathbf{X}_M^{(1{:}T)} \,\right],
\end{equation}
with each $\mathbf{X}_m^{(1{:}T)} \in \mathbb{R}^{H \times W \times T}$ and $[\cdot | \cdot]$ denoting channel-wise concatenation.
The composite tensor $\mathbf{X}$ is partitioned into $P$ non-overlapping spatial patches $\{\mathbf{X}_p\}_{p=1}^P$, where each $\mathbf{X}_p \in \mathbb{R}^{C \times z \times z}$ captures a localized $z \times z$ region. The model predicts the corresponding precipitation patch $Y_p$ for each $\mathbf{X}_p$, and the union $\cup_{p=1}^P Y_p$ reconstructs the full target map $Y^{T+\Delta}$. This patch-wise multimodal representation enables localized, temporally aligned, and cross-modally integrated reasoning for precipitation forecasting.
\begin{table*}[t]
\centering
\scriptsize
\setlength{\tabcolsep}{2pt}
\caption{LPA and Precipitation Categorization for Indian States}
\label{tab:lpa_precip_class}
\begin{tabular}{@{}l l l l l l l l@{}}
\toprule
\textbf{Region} 
  & \makecell{\textbf{Latitude}\\\textbf{Longitude}} 
  & \makecell{\textbf{LPA}\\(mm)}
  & \makecell{\textbf{Scarcity}\\ mm (N\%)} 
  & \makecell{\textbf{Deficit}\\ mm (N\%)} 
  & \makecell{\textbf{Normal}\\ mm (N\%)} 
  & \makecell{\textbf{Excess}\\ mm (N\%)} 
  & \makecell{\textbf{Large} \textbf{Exc.} \\ mm (N\%)} \\
\midrule

Assam             & \makecell{$24^{\circ}13' N–28^{\circ}00' N$\\$89^{\circ}$46' E–$96^{\circ}04' E$}
                  & 328.6  
                  & \makecell{[0–131)\\ 1.88\%}   
                  & \makecell{[131–263)\\ 57.03\%}  
                  & \makecell{[263–394)\\ 40.44\%}
                  & \makecell{[395–526)\\ 0.65\%}  
                  & \makecell{$\ge$527\\ 0.00\%} 
                  \\[10pt]
                  
Bihar             &\makecell{$24^{\circ}20' N–27^{\circ}31' N$\\$83^{\circ}19' E–88^{\circ}17' E$}
                  & 216.5  
                  & \makecell{[0–87)\\ 1.32\%}   
                  & \makecell{[87–173)\\ 30.31\%}  
                  & \makecell{[173–260)\\ 51.58\%}  
                  & \makecell{[260–346)\\ 14.08\%}  
                  & \makecell{$\ge$346 \\ 2.71\%}
                  \\[12pt]
                  
\makecell{Himachal\\ Pradesh}  
                  & \makecell{$30^{\circ}22' N–33^{\circ}12' N$\\$75^{\circ}47' E–79^{\circ}04' E$}
                  & 120.5  
                  & \makecell{[0–48)  \\ 13.13\%}   
                  & \makecell{[48–96) \\ 21.06\%}
                  & \makecell{[96–145) \\ 16.61\%}
                  & \makecell{[145–193) \\ 11.34\%}
                  & \makecell{$\ge$192.8 \\ 37.85\%}
                  \\[12pt]
                  
Karnataka         & \makecell{$11^{\circ}30' N–18^{\circ}30' N$\\$74^{\circ}00' E–78^{\circ}30' E$}
                  & 271.8  
                  & \makecell{[0–109)  \\ 23.60\%}  
                  & \makecell{[109–217)  \\ 39.40\%}
                  & \makecell{[217–326)  \\ 17.67\%}
                  & \makecell{[326–435)  \\ 4.87\%}
                  & \makecell{$\ge$435  \\ 14.47\%}
                  \\[12pt]
                  
Kerala            & \makecell{$8^{\circ}18' N–12^{\circ}48' N$\\$74^{\circ}52' E–77^{\circ}22' E$}
                  & 144.1  
                  & \makecell{[0–58)  \\ 9.47\%} 
                  & \makecell{[58–115)  \\ 51.65\%}  
                  & \makecell{[115–173)  \\ 38.35\%}
                  & \makecell{[173–231)  \\  0.53\%}
                  & \makecell{$\ge$231  \\  0.00\%} \\
\bottomrule
\end{tabular}
\vspace{-0.32cm}
\end{table*}

Each multimodal input patch $\mathbf{X}_p \in \mathbb{R}^{C \times z \times z}$ is given as input to a temporally aware multimodal Attention U-Net architecture \cite{atten_unet}, which extends the traditional U-Net \cite{unet} by integrating multi-source atmospheric inputs across time and applying spatial attention gates to focus on precipitation-relevant features.  A schematic overview of the proposed model is illustrated in Figure~\ref{fig:mvforecast}. The base architecture consists of a shared convolutional encoder and decoder with attention-guided skip connections. Each input patch undergoes hierarchical feature extraction via successive convolution, batch normalization, rectilinear unit activations, and max pooling layers, as shown in  Figure~\ref{fig:mvforecast}. The attention gates compute spatial relevance maps at each decoder level, selectively emphasizing salient regions in the encoder feature maps. These architectural components are inherited from the Attention U-Net model.

\begin{figure*}[t]
  \centering
  \begin{subfigure}[b]{0.235\textwidth}
    \includegraphics[width=\textwidth, trim=20 20 6 18, clip]{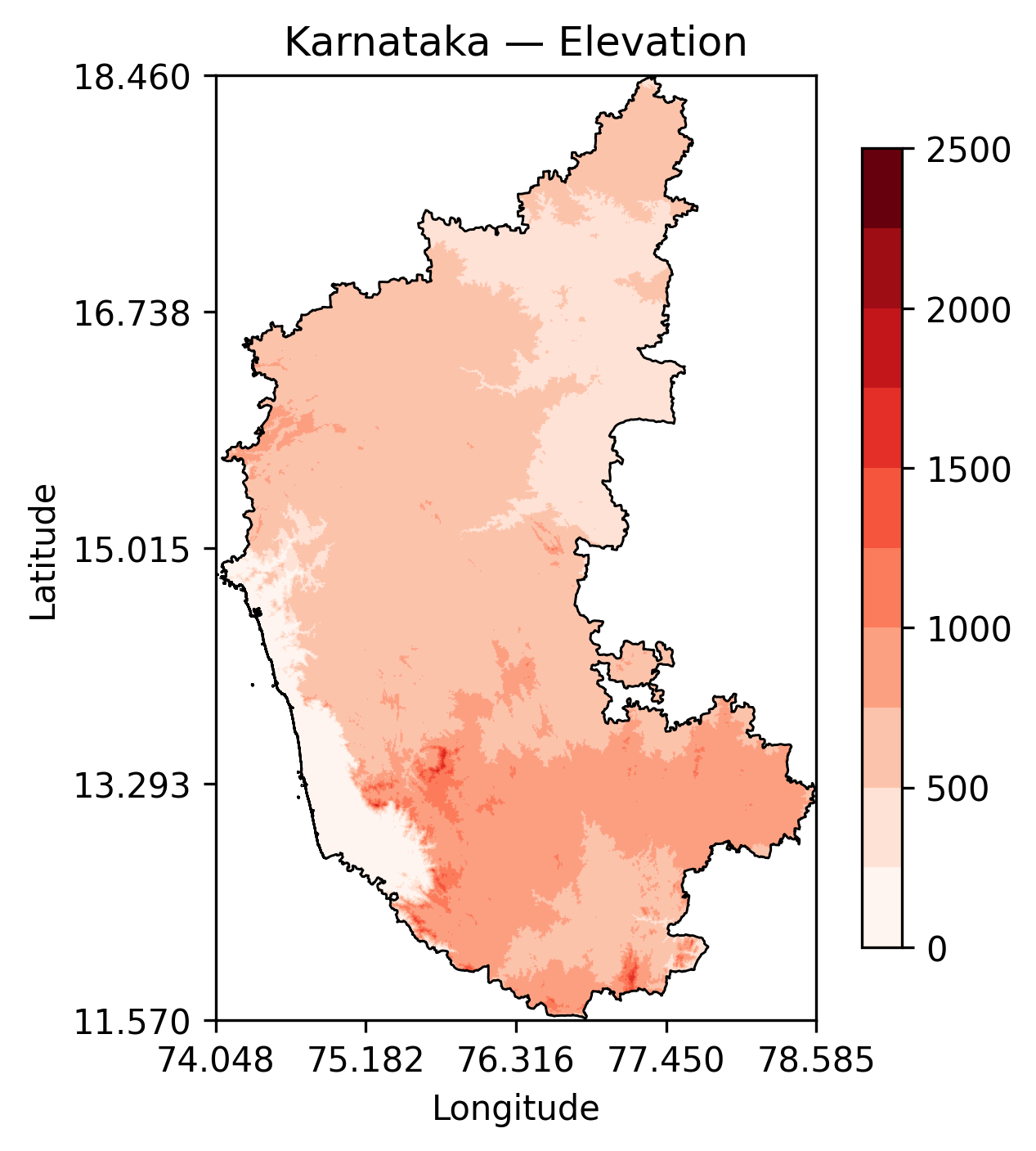}
    \caption{\scriptsize Elevation}
  \end{subfigure}
  \hspace{5pt}
  \begin{subfigure}[b]{0.22\textwidth}
    \includegraphics[width=\textwidth, trim=20 20 6 18, clip]{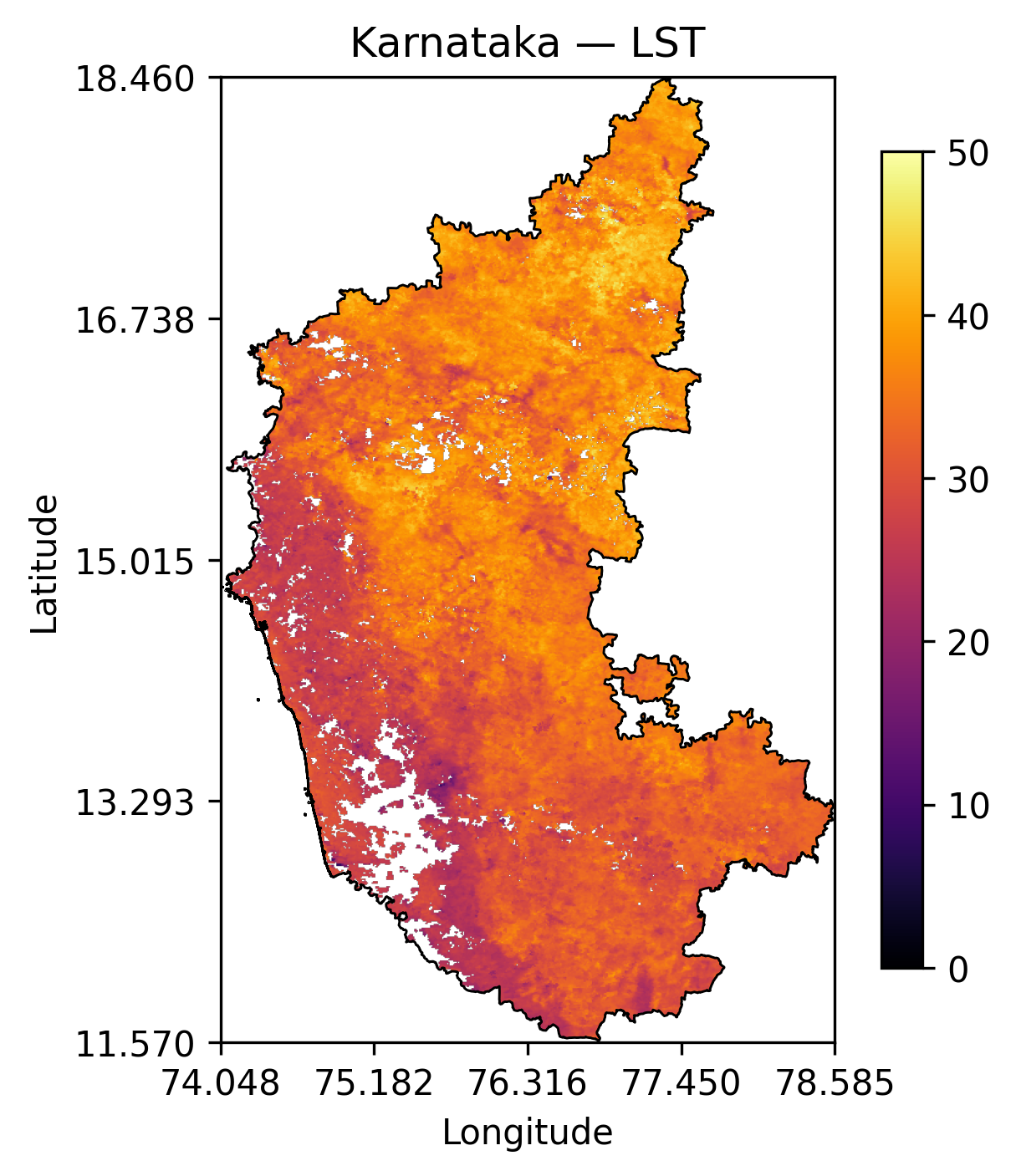}
    \caption{\scriptsize LST}
  \end{subfigure}
  \hspace{5pt}
  \begin{subfigure}[b]{0.215\textwidth}
    \includegraphics[width=\textwidth, trim=20 20 6 18, clip]{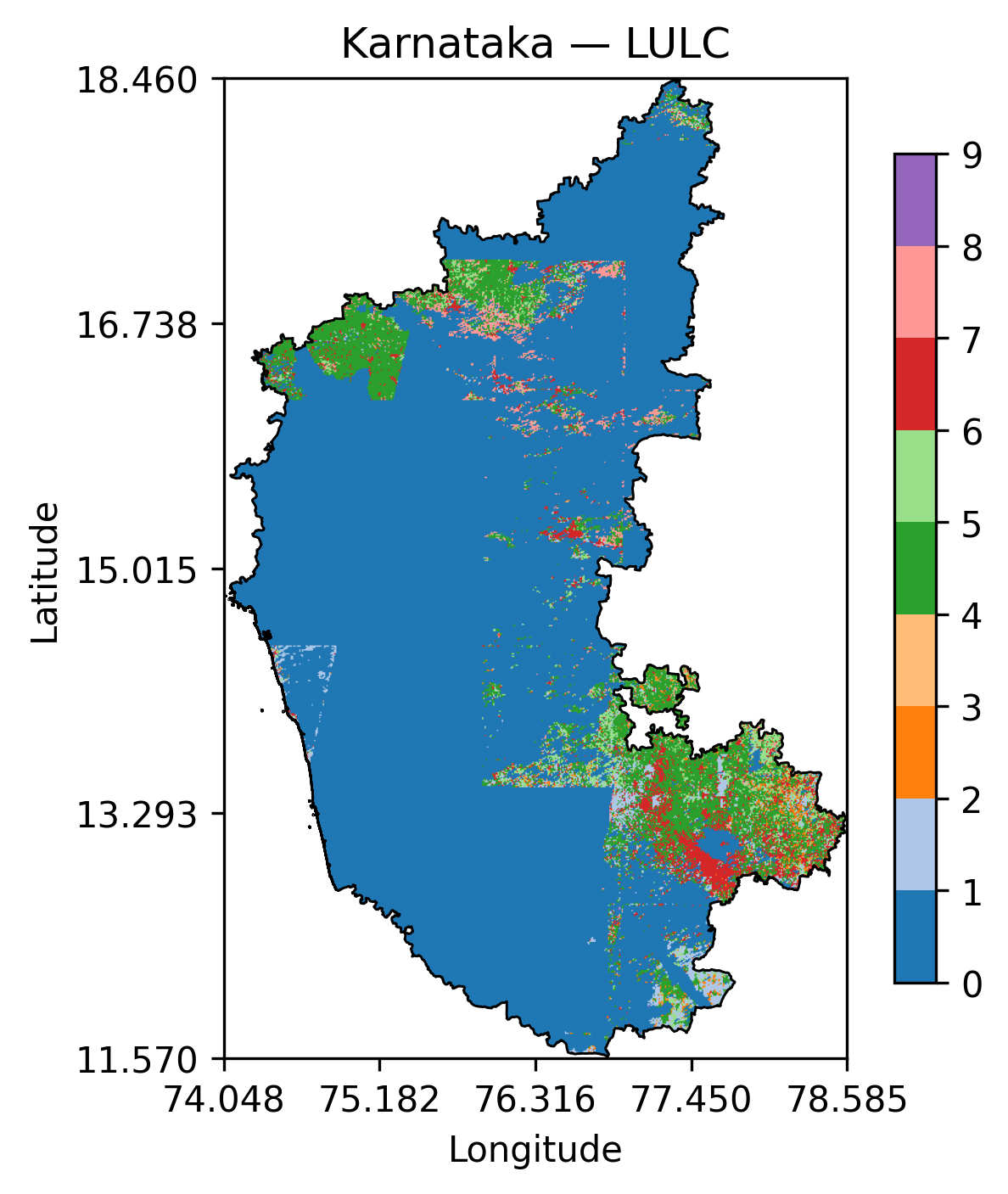}
    \caption{\scriptsize LULC}
  \end{subfigure}
  \hspace{5pt}
  \begin{subfigure}[b]{0.23\textwidth}
    \includegraphics[width=\textwidth, trim=20 20 6 18, clip]{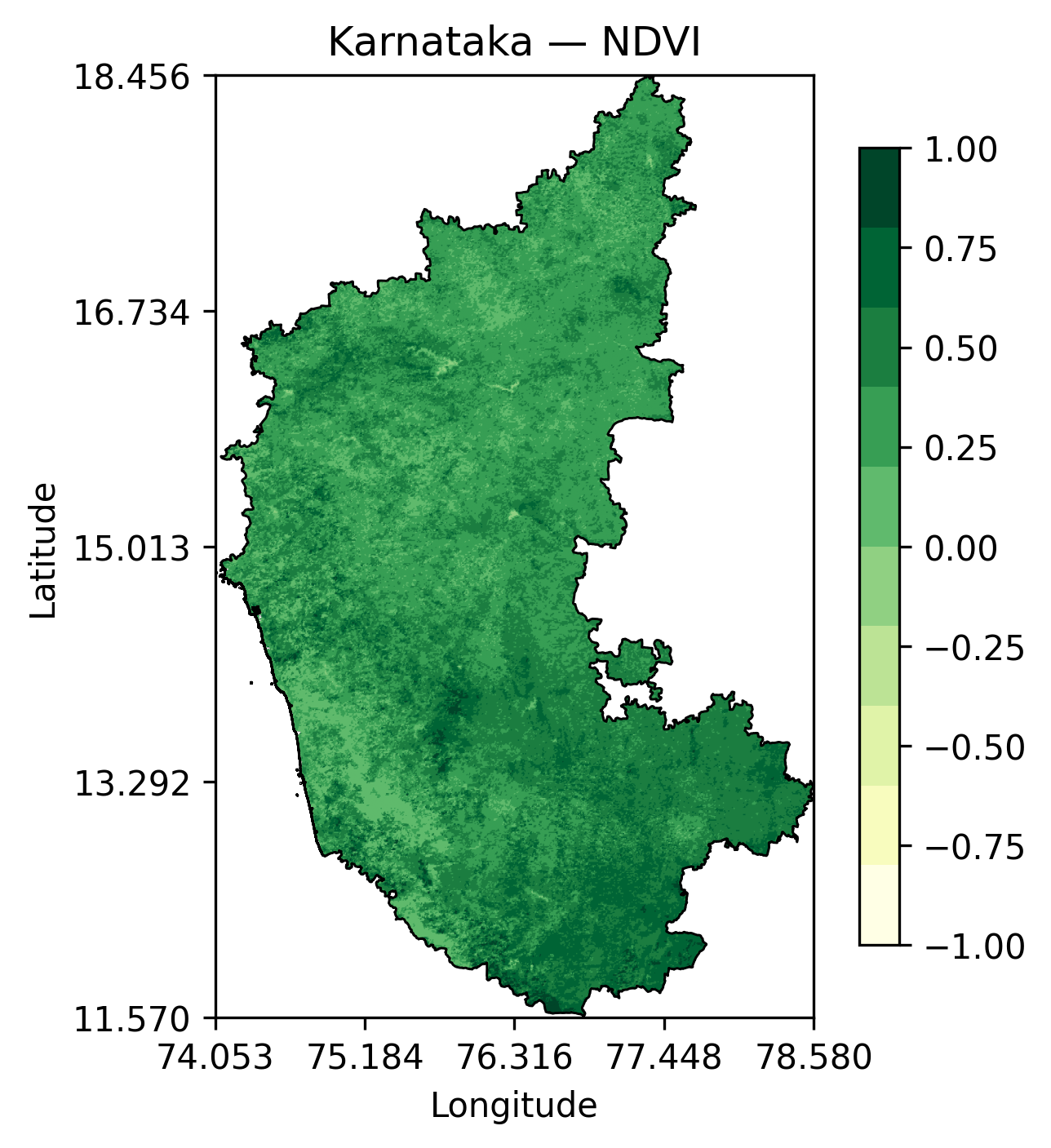}
    \caption{\scriptsize NDVI}
  \end{subfigure}

  \begin{subfigure}[b]{0.225\textwidth}
    \includegraphics[width=\textwidth, trim=20 20 6 18, clip]{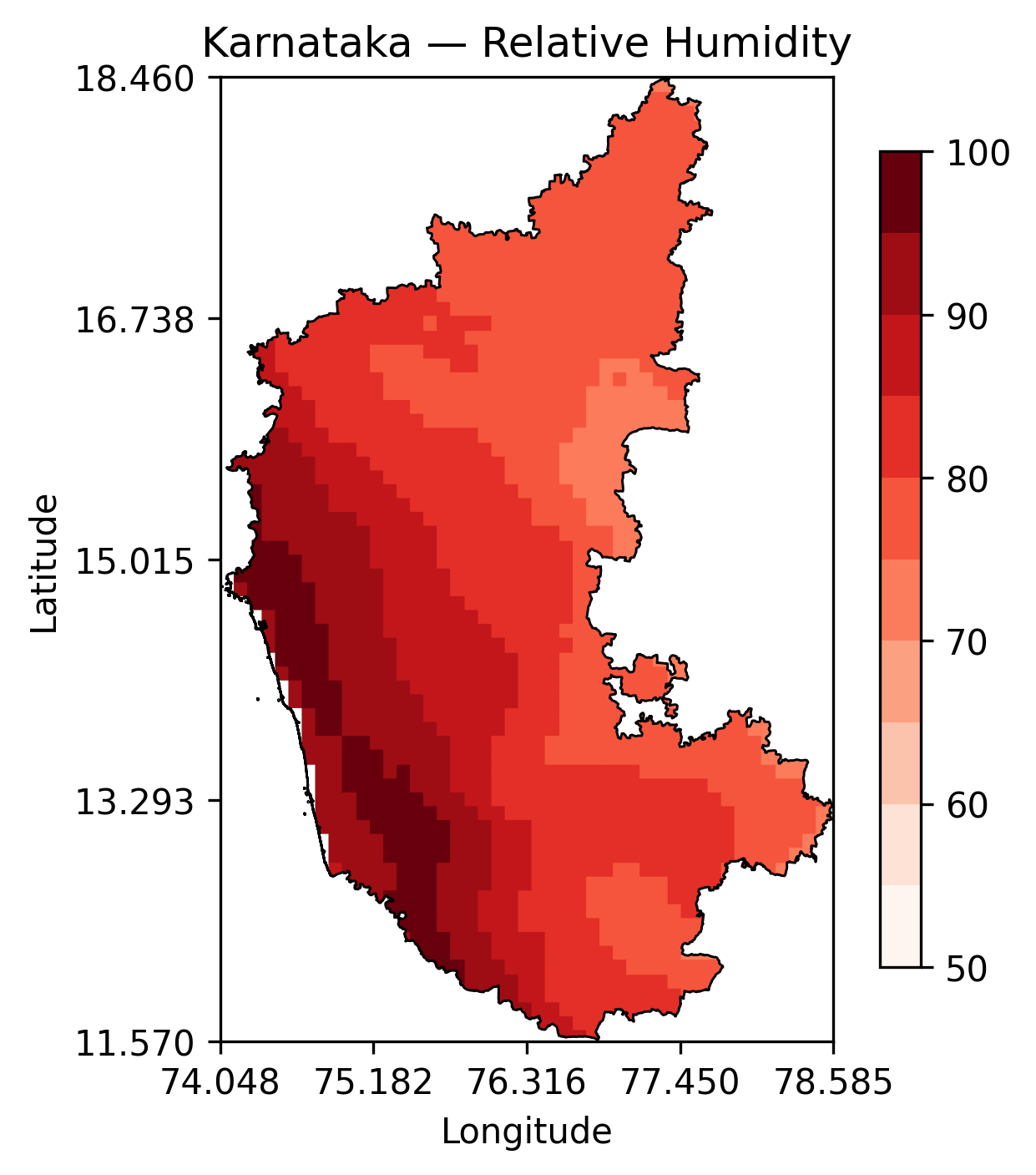}
    \caption{\scriptsize Humidity}
  \end{subfigure}
  \hspace{5pt}
  \begin{subfigure}[b]{0.22\textwidth}
    \includegraphics[width=\textwidth, trim=20 20 6 18, clip]{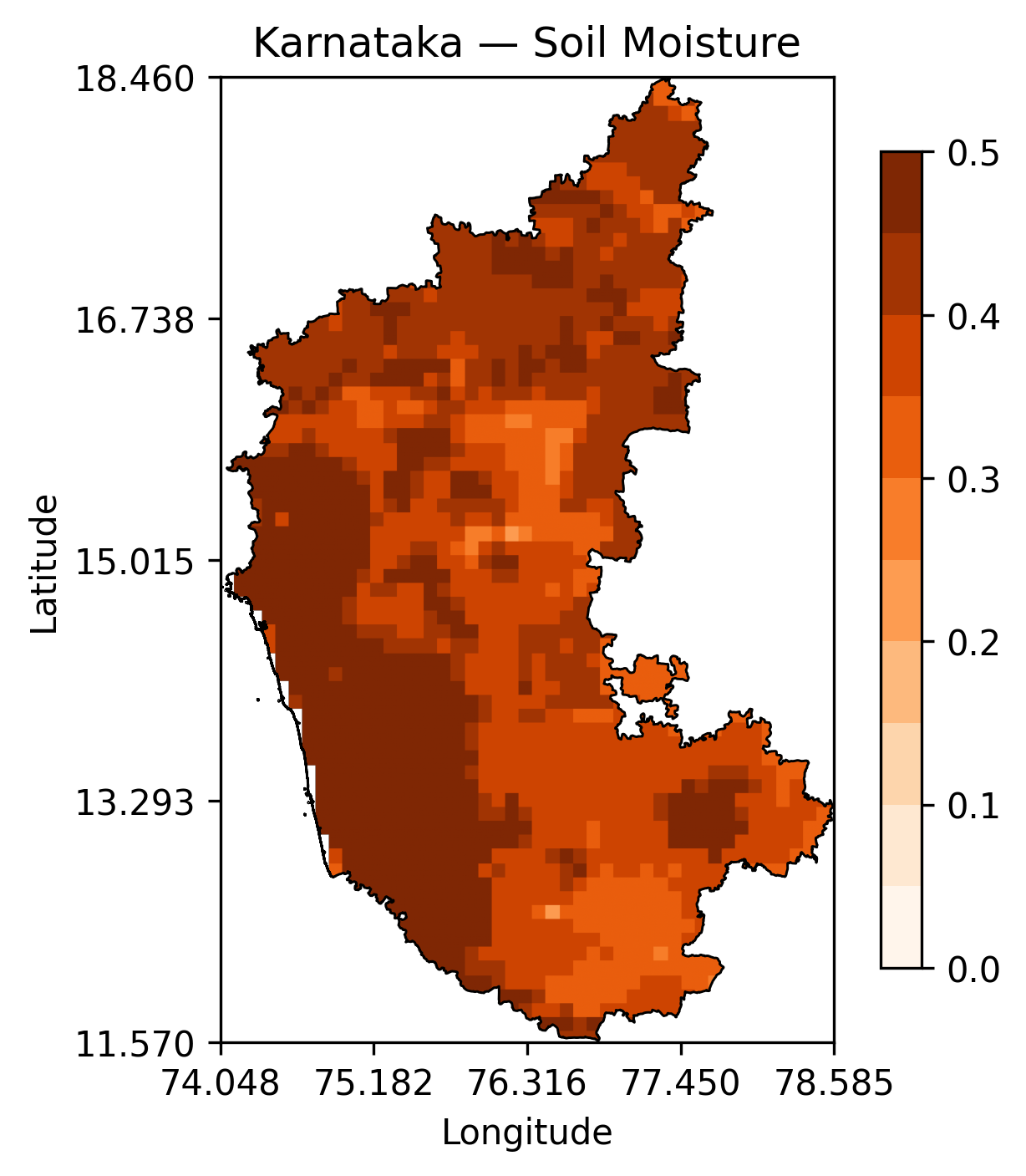}
    \caption{\scriptsize Soil Moisture}
  \end{subfigure}
  \hspace{4pt}
  \begin{subfigure}[b]{0.22\textwidth}
    \includegraphics[width=\textwidth, trim=20 20 6 18, clip]{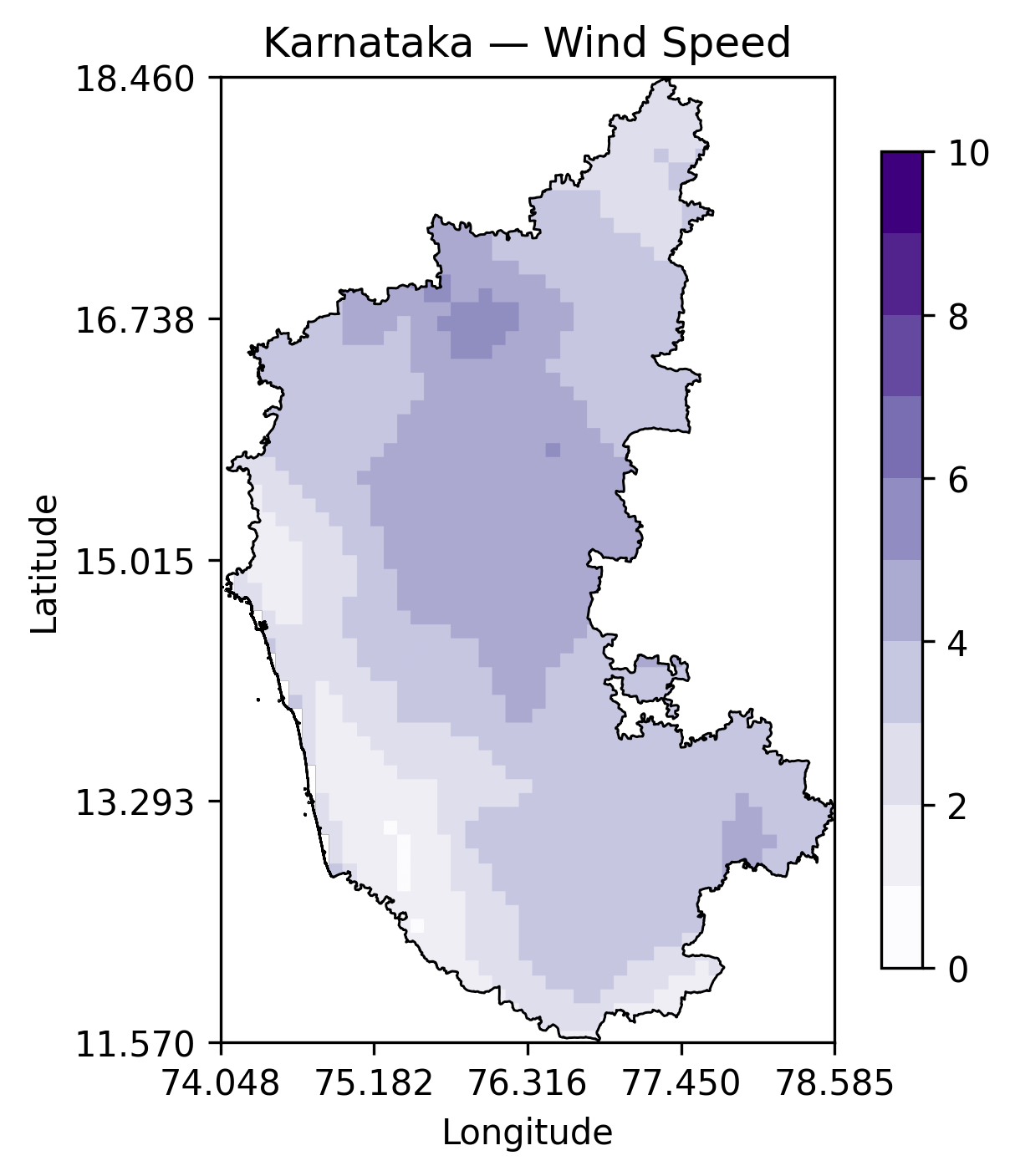}
    \caption{\scriptsize Wind Speed}
  \end{subfigure}
  \hspace{0.5pt}
  \begin{subfigure}[b]{0.23\textwidth}
    \includegraphics[width=\textwidth, trim=20 20 6 18, clip]{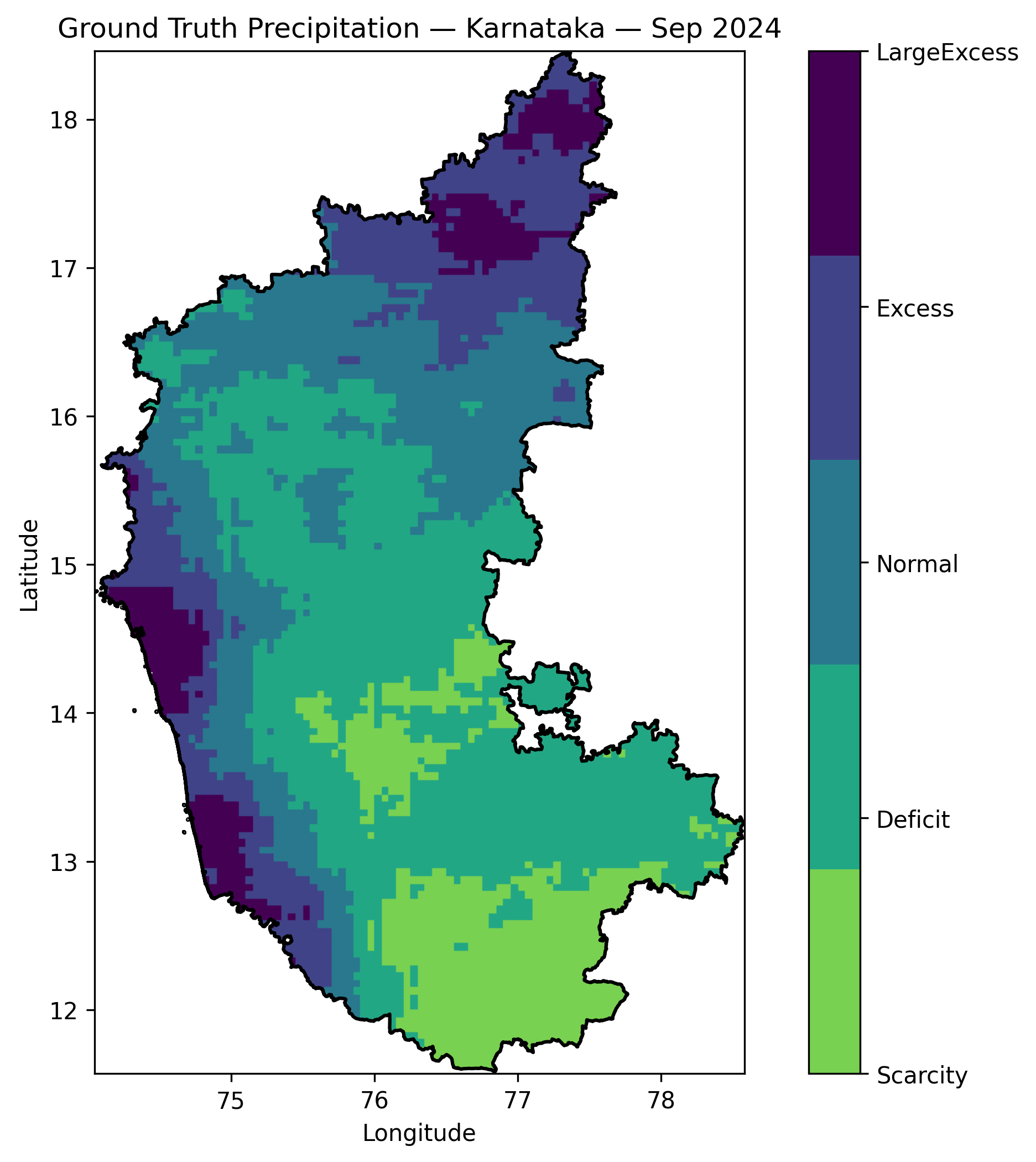}
    \caption{\scriptsize Precipitation}
  \end{subfigure}

  \caption{Multimodal inputs for June and precipitation target for September 2024 for Karnataka, India.}
  \label{fig:Karnataka_views}
\vspace{-0.3cm}
\end{figure*}

\subsection{Loss Function and Training}
To train the model, a composite loss is employed that balances grid-point-wise classification accuracy with spatial consistency. The \textit{Focal Loss} \cite{focal_loss} focuses on hard-to-classify grid points by down-weighting confident predictions. For a ground-truth class \(k\) at location \((i,j)\), it is defined as:
\begin{equation}
\mathcal{L}_{\mathrm{FL}} = \frac{1}{|Y_p|} \sum_{(i,j) \in Y_p} 
\alpha \left(1 - \hat{Y}_{i,j}^{(k)}\right)^\gamma \left(-\log \hat{Y}_{i,j}^{(k)}\right),
\end{equation}
where \(\alpha\) and \(\gamma\) control the weighting on misclassified points.

To promote region-level agreement, the \textit{Dice Loss} \cite{dice_loss} is used. For each class \(k \in \{1, \dots, K\}\), it is defined as:
\begin{equation}
\mathrm{Dice}_k = \frac{2\sum_{i,j} \hat{Y}_{i,j}^{(k)} Y_{i,j}^{(k)} + \epsilon}
{\sum_{i,j} \hat{Y}_{i,j}^{(k)} + \sum_{i,j} Y_{i,j}^{(k)} + \epsilon}, \quad \text{ and \space }
\mathcal{L}_{\mathrm{Dice}} = 1 - \frac{1}{K} \sum_{k=1}^K \mathrm{Dice}_k,
\end{equation}
with \(Y_{i,j}^{(k)}\) as the one-hot label and \(\epsilon = 1.0\) for stability.

 The overall loss is formulated as a weighted sum of both terms, with relative weighting hyperparameters $\lambda_{\mathrm{FL}}$ and $\lambda_{\mathrm{Dice}}$.
\begin{equation}
\mathcal{L} = \lambda_{\mathrm{FL}} \mathcal{L}_{\mathrm{FL}} + \lambda_{\mathrm{Dice}} \mathcal{L}_{\mathrm{Dice}},
\end{equation}
encouraging class-discriminative yet spatially coherent predictions essential for reliable precipitation classification.

\section{Results}
This section presents the results of the proposed multimodal precipitation forecasting model. We begin with an overview of the data curation process, followed by the experimental setup for the proposed and baseline methods. Next, quantitative and qualitative evaluations are used to demonstrate the model’s effectiveness in capturing spatially distributed precipitation patterns.

\begin{figure*}[t]
\vspace{-0.3cm}
  \centering
  \begin{subfigure}[b]{0.23\textwidth}
    \includegraphics[width=\textwidth, trim=20 20 6 18, clip]{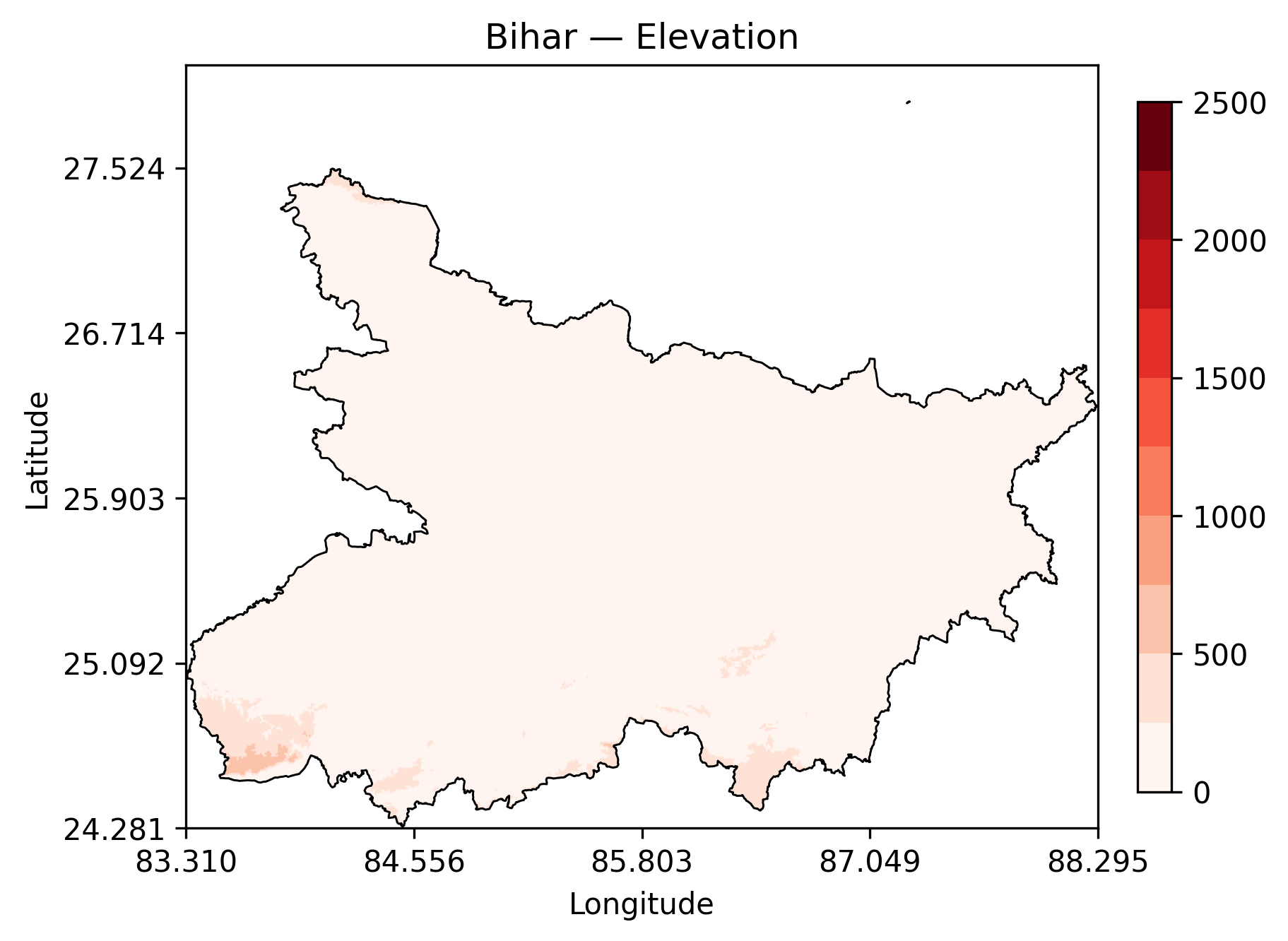}
    \caption{\scriptsize Elevation}
  \end{subfigure}
  \hspace{4.6pt}
  \begin{subfigure}[b]{0.225\textwidth}
    \includegraphics[width=\textwidth, trim=20 20 6 18, clip]{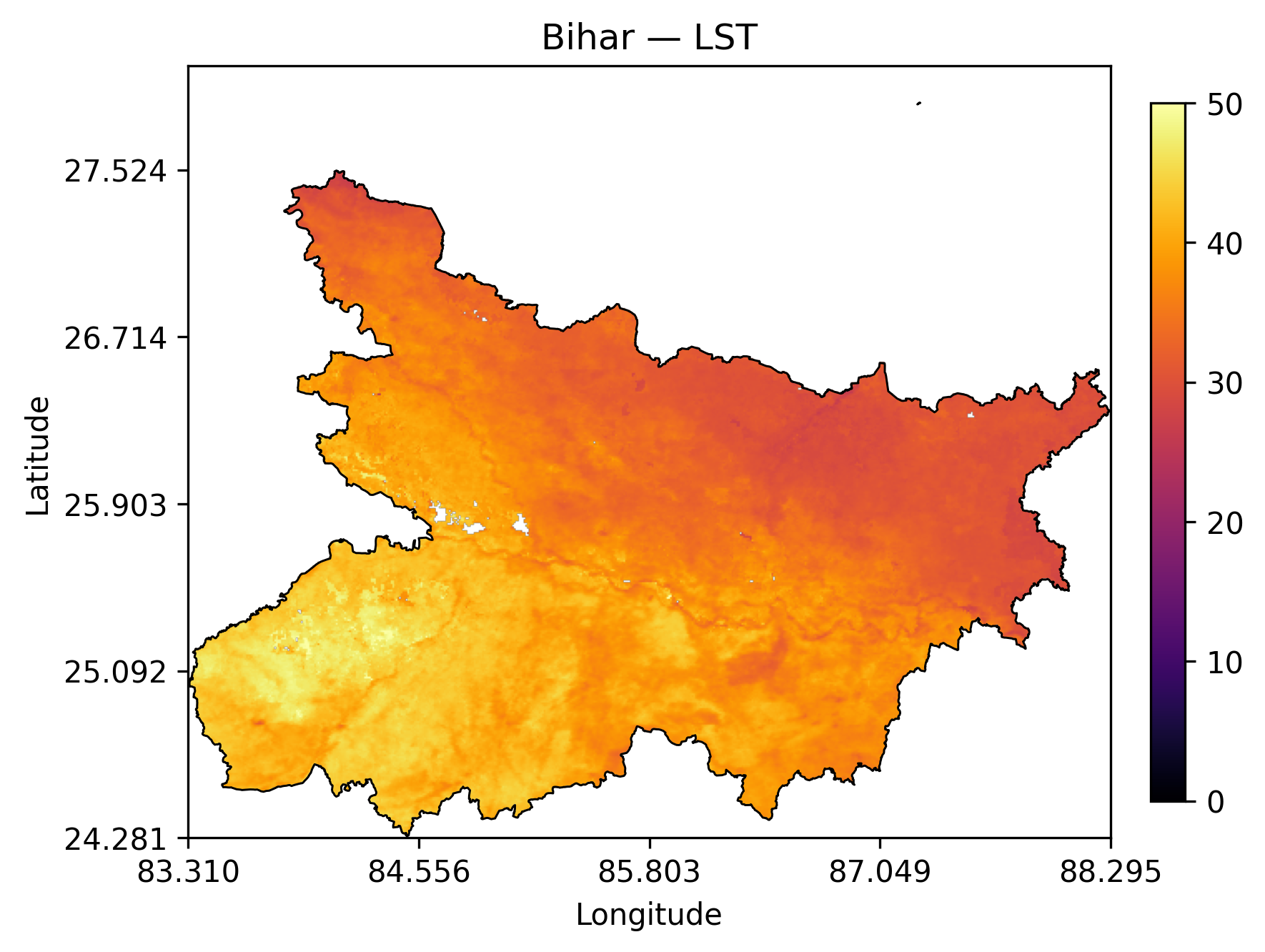}
    \caption{\scriptsize LST}
  \end{subfigure}
  \hspace{5pt}
  \begin{subfigure}[b]{0.22\textwidth}
    \includegraphics[width=\textwidth, trim=20 20 6 18, clip]{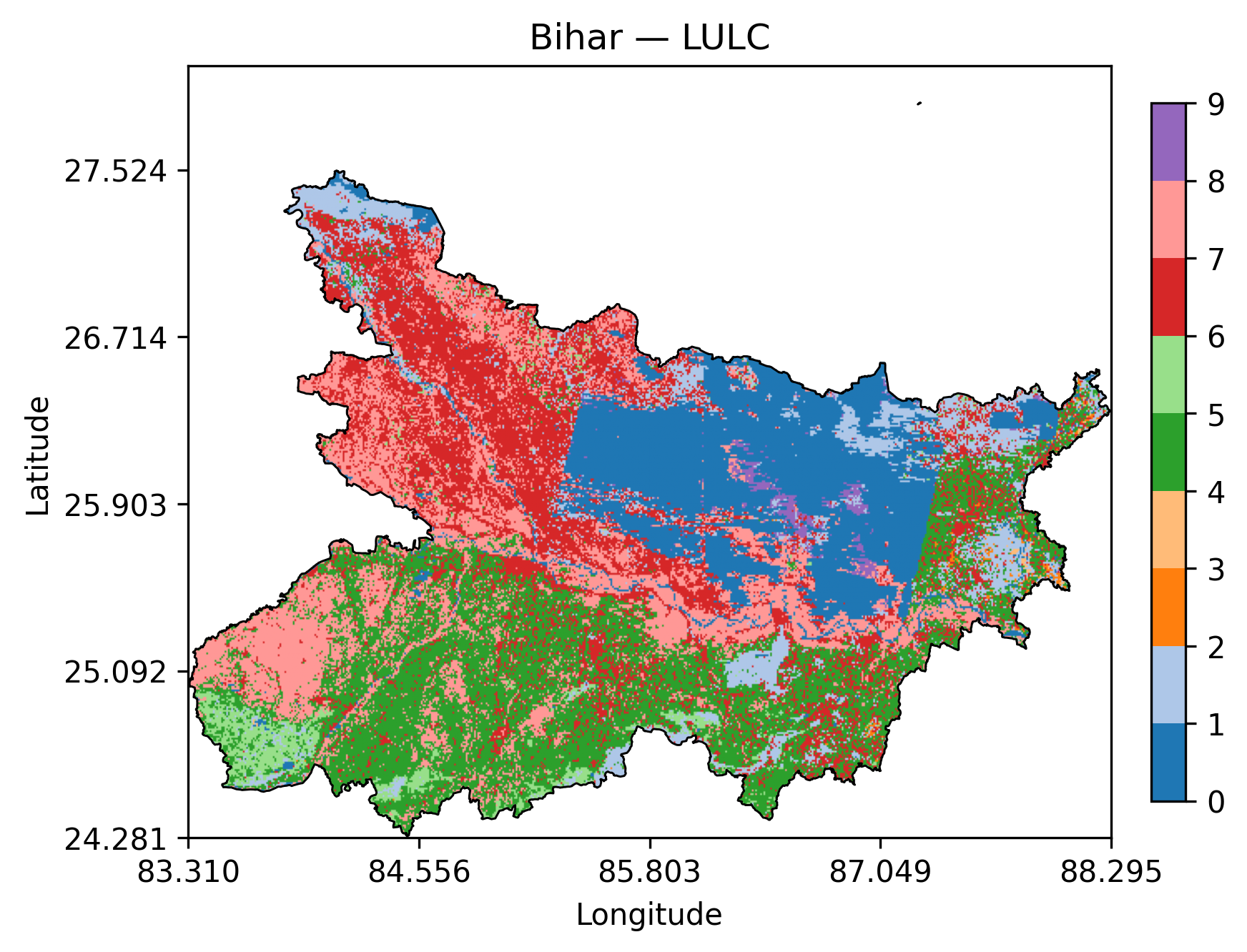}
    \caption{\scriptsize LULC}
  \end{subfigure}
  \hspace{5pt}
  \begin{subfigure}[b]{0.236\textwidth}
    \includegraphics[width=\textwidth, trim=20 20 6 18, clip]{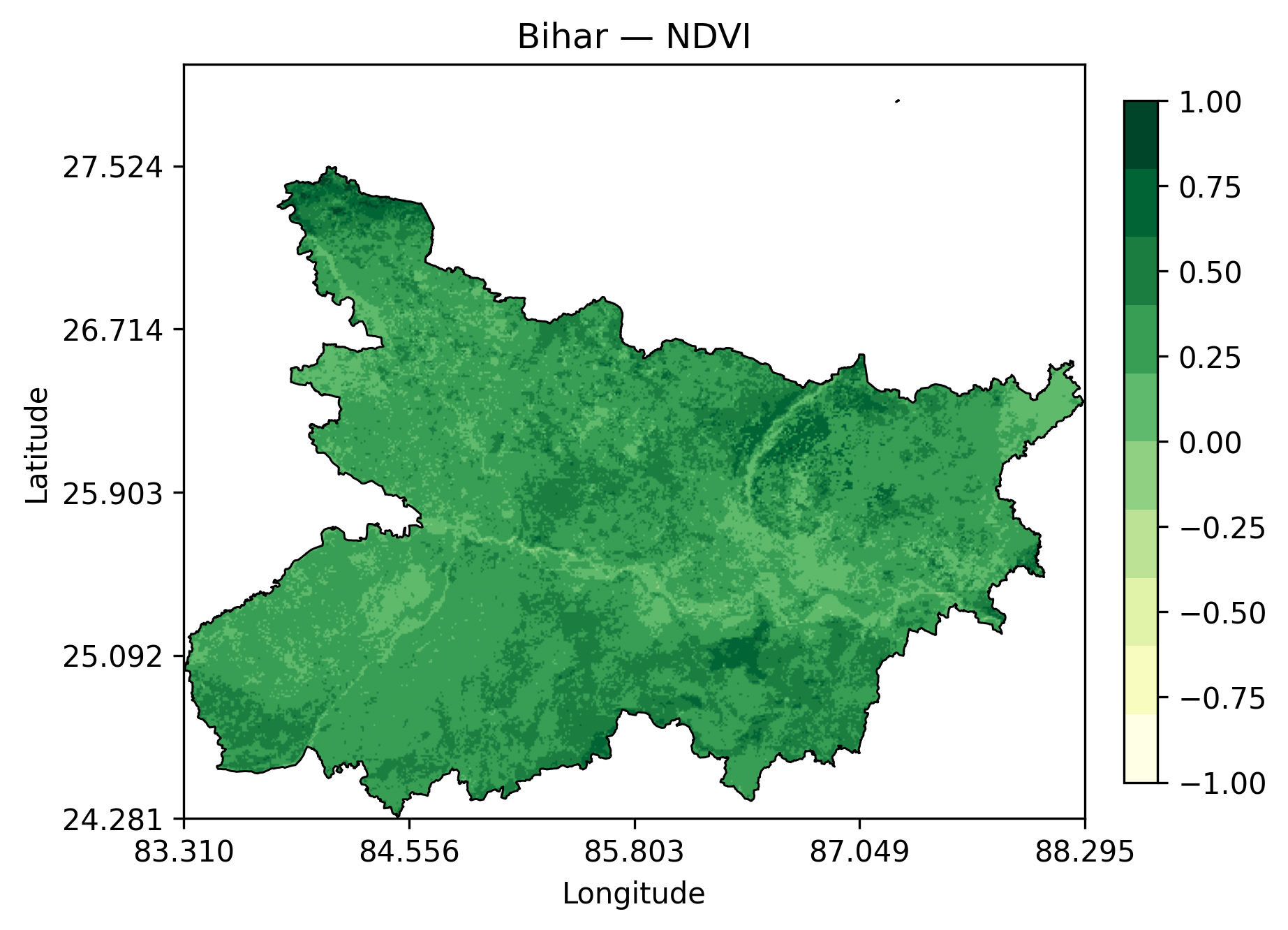}
    \caption{\scriptsize NDVI}
  \end{subfigure}

  \begin{subfigure}[b]{0.225\textwidth}
    \includegraphics[width=\textwidth, trim=20 20 6 18, clip]{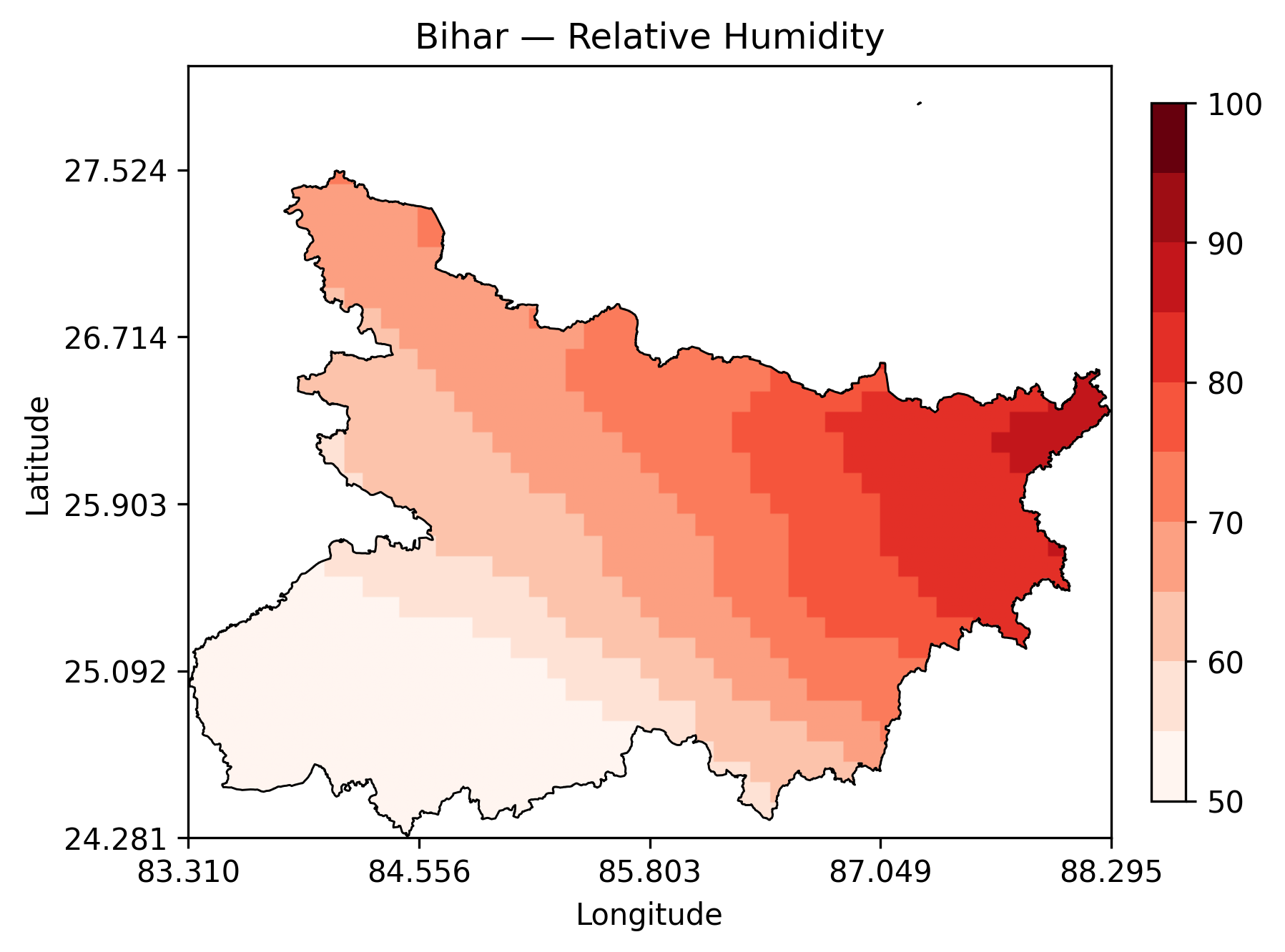}
    \caption{\scriptsize Humidity}
  \end{subfigure}
  \hspace{5pt}
  \begin{subfigure}[b]{0.225\textwidth}
    \includegraphics[width=\textwidth, trim=20 20 6 18, clip]{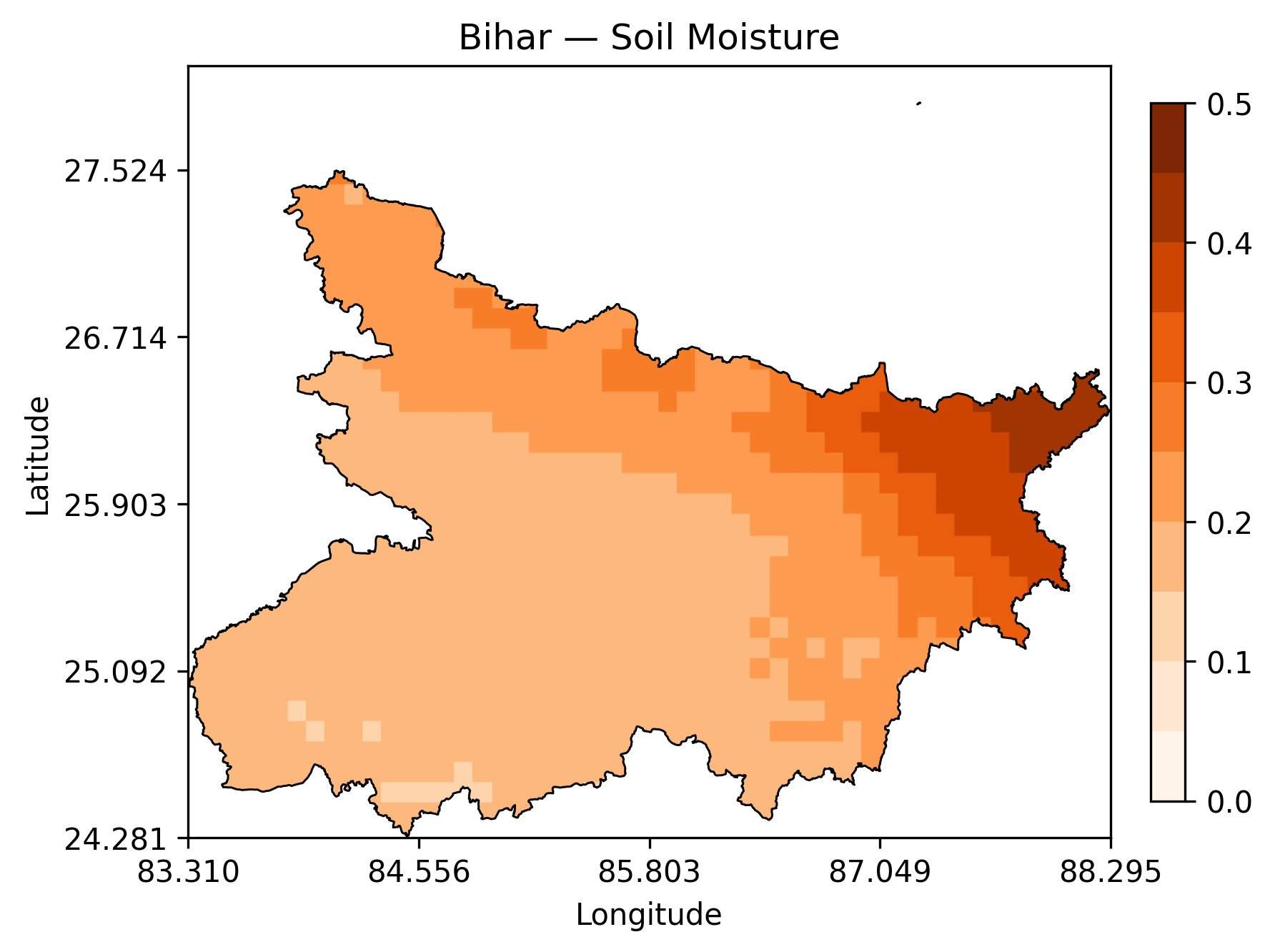}
    \caption{\scriptsize Soil Moisture}
  \end{subfigure}
  \hspace{4pt}
  \begin{subfigure}[b]{0.22\textwidth}
    \includegraphics[width=\textwidth, trim=20 20 6 18, clip]{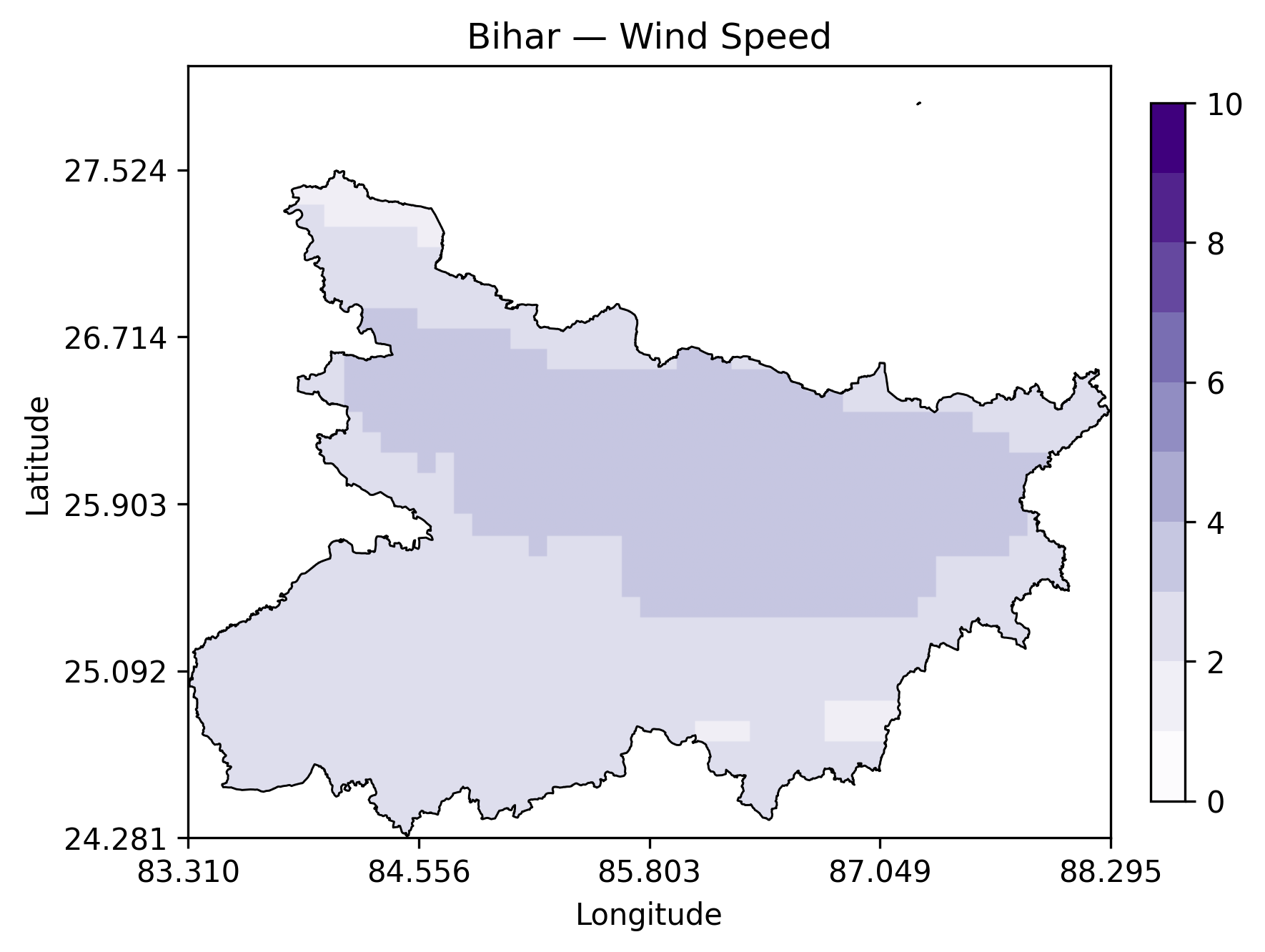}
    \caption{\scriptsize Wind Speed}
  \end{subfigure}
  \hspace{0.5pt}
  \begin{subfigure}[b]{0.25\textwidth}
    \includegraphics[width=\textwidth, trim=20 20 6 18, clip]{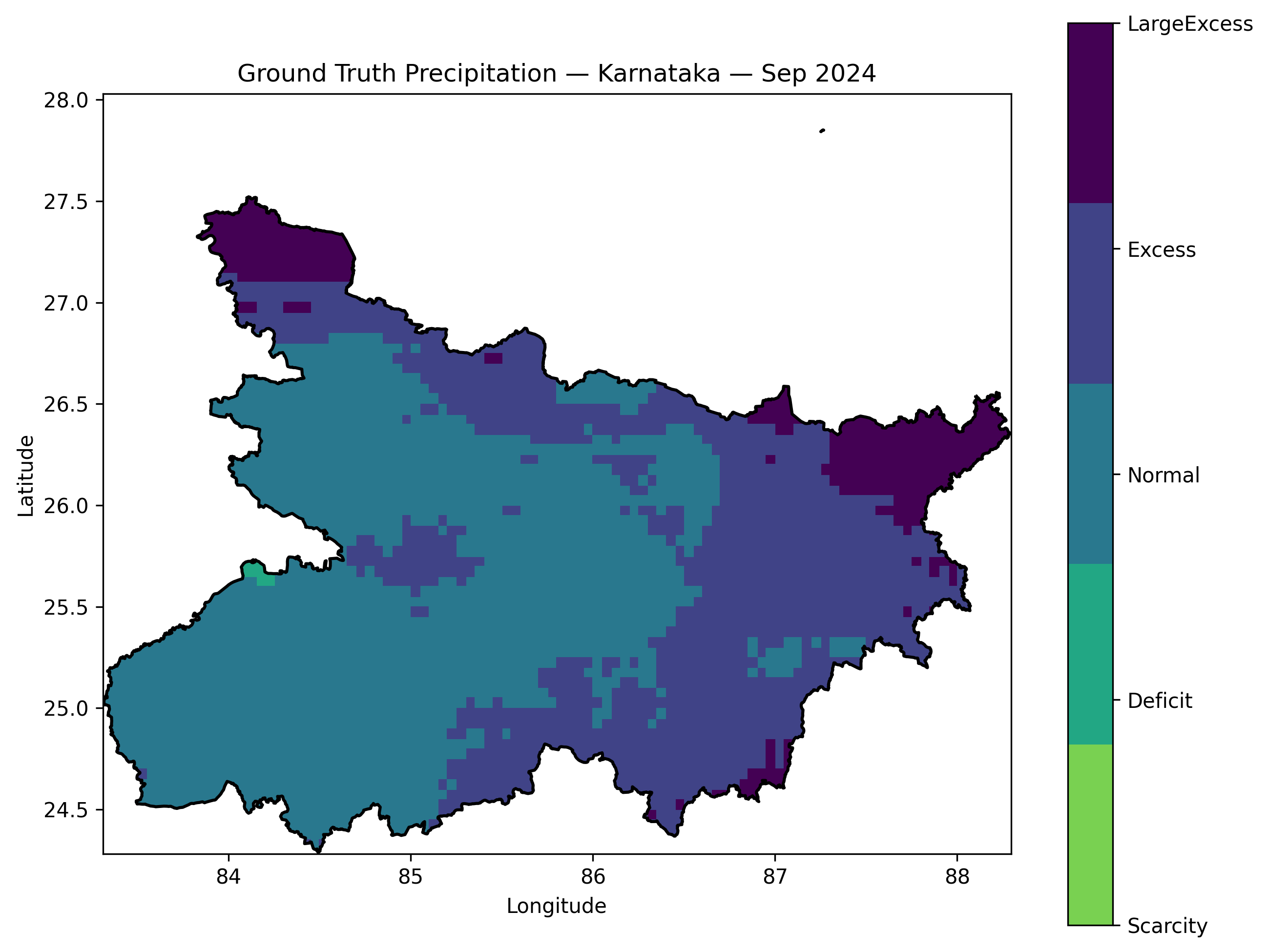}
    \caption{\scriptsize Precipitation}
  \end{subfigure}
  \caption{Multimodal inputs for June and precipitation target for September 2024 for Bihar, India.}
  \label{fig:Bihar_views}
  \vspace{-0.4cm}
\end{figure*}

\subsection{Multimodal Precipitation Data Curation}
To evaluate the performance of the proposed model in the prediction of high-resolution precipitation on the Indian subcontinent, a comprehensive spatially aligned and temporally indexed multimodal data set was curated using satellite-derived Earth observations via Google Earth Engine\footnote{\url{http://earthengine.google.com}}. All input variables were resampled and harmonized to a uniform grid resolution of 1 km × 1 km to enable fine-scale regional forecasting. This data is of higher resolution compared to the standard climate variables like soil moisture, wind speed, and precipitation which are typically available only at coarse resolutions ranging from 5 km to 50 km in satellite based earth observation datasets such as ERA5\footnote{\url{https://www.ecmwf.int/en/forecasts/dataset/ecmwf-reanalysis-v5}}, and CHIRPS\footnote{\url{https://www.chc.ucsb.edu/data/chirps}}. Our dataset spans five representative Indian states of Assam, Bihar, Himachal Pradesh, Kerala, and Karnataka for the 2024 monsoon season. Predictor features were extracted for $T = 3$ months (June–August), with the precipitation forecast target set at a lead time of $\Delta = 1$ month (September).

\begin{table*}[t]
\vspace{-0.5cm}
\centering
\small\caption{Ablation Study: Effect of Individual Modalities vs. Multimodal Fusion}
\label{tab:ind_vs_multi}
\scriptsize
\renewcommand{\arraystretch}{0.7}
\setlength{\tabcolsep}{2pt}
\setcellgapes{1pt}\makegapedcells
\begin{threeparttable}
\begin{tabular}{@{}llcc
  @{\hspace{3pt}}c
  @{\hspace{3pt}}c
  @{\hspace{3pt}}c
  @{\hspace{3pt}}c
  @{\hspace{3pt}}c@{}}
\toprule
\multirow{2}{*}{\textbf{State}} & \multirow{2}{*}{\textbf{Model}} &
  \multirow{2}{*}{\makecell{\textbf{Overall}\\\textbf{Accuracy}}} &
  \multirow{2}{*}{\makecell{\textbf{Weighted}\\\textbf{F1 Score}}} &
  \multicolumn{5}{c}{\textbf{Class-wise F1 Score}} \\
\cmidrule(lr){5-9}
  &  &  &  & \textbf{Scarcity} & \textbf{Deficit}
         & \textbf{Normal}  & \textbf{Excess}  & \textbf{Large Excess} \\
\midrule

\multirow{8}{*}{Assam}
  & Elevation   & 0.4668 & 0.4223 & 0.2509 & 0.6580 & 0.0596 & 0.0000 & -- \\
  & LST         & 0.5305 & 0.5033 & 0.4982 & 0.4351 & 0.6304 & 0.0000 & -- \\
  & LULC        & 0.5779 & 0.5910 & 0.3604 & 0.7079 & 0.4759 & 0.0000 & -- \\
  & NDVI        & 0.3314 & 0.3739 & 0.4693 & 0.3438 & 0.2509 & --     & -- \\
  & Humidity    & 0.6838 & 0.6863 & 0.8953 & 0.7553 & 0.2538 & 0.0000 & -- \\
  & Soil Moist. & 0.4810 & 0.4406 & 0.3482 & 0.6038 & 0.1211 & 0.0000 & -- \\
  & Wind Speed  & 0.6830 & 0.6841 & 0.7276 & 0.7674 & 0.4063 & 0.0000 & -- \\
  & \textbf{Proposed}  & \textbf{0.9045} & \textbf{0.9038}
                 & \textbf{0.9245} & \textbf{0.8932}
                 & \textbf{0.8996} & \textbf{1.0000} & -- \\
\midrule

\multirow{8}{*}{Bihar}
  & Elevation   & 0.4821 & 0.4870 & --      & 0.0000 & 0.6261 & 0.4081 & 0.0414 \\
  & LST         & 0.7772 & 0.7827 & --      & 0.0000 & 0.8559 & 0.7081 & 0.5352 \\
  & LULC        & 0.5933 & 0.5664 & --      & --     & 0.7012 & 0.4810 & 0.3312 \\
  & NDVI        & 0.6288 & 0.5892 & --      & --     & 0.7410 & 0.5082 & 0.0856 \\
  & Humidity    & 0.7919 & 0.7852 & --      & --     & 0.8668 & 0.6094 & 0.8111 \\
  & Soil Moist. & 0.6829 & 0.6844 & --      & --     & 0.7695 & 0.6258 & 0.3187 \\
  & Wind Speed  & 0.8051 & 0.7948 & --      & --     & 0.8847 & 0.7261 & 0.1438 \\
  & \textbf{Proposed}   & \textbf{0.9322} & \textbf{0.9302}
                 & --      & 0.0000 & \textbf{0.9153}
                 & \textbf{0.9335} & \textbf{0.9758} \\
\midrule

\multirow{8}{*}{Himachal}
  & Elevation   & 0.4705 & 0.4203 & 0.0729 & 0.2888 & 0.4301 & 0.1487 & 0.6855 \\
  & LST         & 0.4887 & 0.4572 & 0.1690 & 0.4582 & 0.0544 & 0.3761 & 0.7735 \\
  & LULC        & 0.4066 & 0.3884 & 0.3577 & 0.2741 & 0.3146 & 0.2785 & 0.5962 \\
  & NDVI        & 0.4914 & 0.4674 & 0.0427 & 0.5545 & 0.3005 & 0.2413 & 0.6547 \\
  & Humidity    & 0.6501 & 0.6419 & 0.6547 & 0.6631 & 0.5861 & 0.3456 & 0.8102 \\
  & Soil Moist. & 0.5466 & 0.5398 & 0.2752 & 0.4183 & 0.5456 & 0.3706 & 0.7473 \\
  & Wind Speed  & 0.5931 & 0.5923 & 0.6832 & 0.6626 & 0.4339 & 0.3824 & 0.7343 \\
  & \textbf{Proposed}   & \textbf{0.8802} & \textbf{0.8837}
                 & \textbf{0.9612} & \textbf{0.9681}
                 & \textbf{0.8690} & \textbf{0.6510}
                 & \textbf{0.8506} \\
\midrule

\multirow{8}{*}{Karnataka}
  & Elevation   & 0.4820 & 0.4545 & 0.3051 & 0.6592 & 0.3055 & 0.4367 & 0.1796 \\
  & LST         & 0.4454 & 0.4303 & 0.4521 & 0.5494 & 0.3425 & 0.3874 & 0.1526 \\
  & LULC        & 0.4810 & 0.4378 & 0.4521 & 0.6018 & 0.2955 & 0.2834 & 0.1692 \\
  & NDVI        & 0.5055 & 0.4873 & 0.5482 & 0.5439 & 0.5322 & 0.3943 & 0.0338 \\
  & Humidity    & 0.6520 & 0.6222 & 0.7525 & 0.8085 & 0.5913 & 0.4834 & 0.0352 \\
  & Soil Moist. & 0.5633 & 0.5719 & 0.6319 & 0.7173 & 0.4490 & 0.3958 & 0.3435 \\
  & Wind Speed  & 0.6464 & 0.6209 & 0.6417 & 0.7101 & 0.6166 & 0.6681 & 0.0000 \\
  & \textbf{Proposed}   & \textbf{0.9027} & \textbf{0.9036}
                 & \textbf{0.9804} & \textbf{0.8219}
                 & \textbf{0.8619} & \textbf{0.8730}
                 & \textbf{0.9363} \\
\midrule

\multirow{8}{*}{Kerala}
  & Elevation   & 0.4633 & 0.4590 & 0.4703 & 0.4653 & 0.0000 & -- & -- \\
  & LST         & 0.5749 & 0.5377 & \textbf{0.7200} & 0.4422 & 0.0000 & -- & -- \\
  & LULC        & 0.4143 & 0.4150 & 0.4294 & 0.4049 & 0.0000 & -- & -- \\
  & NDVI        & 0.4086 & 0.3974 & 0.3820 & 0.4949 & 0.0077 & -- & -- \\
  & Humidity    & 0.6771 & 0.6284 & 0.5008 & \textbf{0.8253} & 0.0000 & -- & -- \\
  & Soil Moist. & 0.6422 & 0.6106 & 0.3864 & 0.7511 & --      & -- & -- \\
  & Wind Speed  & 0.7121 & 0.6864 & 0.7110 & 0.7703 & 0.2373 & -- & -- \\
  & \textbf{Proposed}   & \textbf{0.7764} & \textbf{0.7649}
                 & 0.6561 & 0.7869
                 & \textbf{0.9527} & -- & -- \\
\bottomrule
\end{tabular}
\begin{tablenotes}
\item $-$ indicates absence of that specific precipitation class in a state.
\end{tablenotes}
\end{threeparttable}
\vspace{-0.5cm}
\end{table*}

The target variable \(Y\) represents the September 2024 categorical precipitation intensity, derived from the CHIRPS dataset. Table~\ref{tab:lpa_precip_class} summarizes the region-wise Long Period Average (LPA) values and quantized precipitation thresholds for each class. These thresholds are derived following the IMD guidelines, customized for each state’s climatology. For instance, Assam has an LPA of 328.6 mm for the monsoon period, with scarcity, deficit, normal, excess, and large excess categories defined as specific rainfall ranges relative to the LPA. This state-wise quantization accounts for the heterogeneity in regional rainfall distribution patterns across India.

To model the drivers of monsoon precipitation, seven geospatial predictors were extracted to capture physical, ecological, and atmospheric processes were extracted from different satellites:
(i) Land Surface Temperature (LST) from MODIS Terra and Aqua sensors 
(\href{https://developers.google.com/earth-engine/datasets/catalog/MODIS_061_MOD11A1}{MOD11A1} and 
\href{https://developers.google.com/earth-engine/datasets/catalog/MODIS_061_MYD11A1}{MYD11A1}); 
(ii) Normalized Difference Vegetation Index (NDVI) from \href{https://developers.google.com/earth-engine/datasets/catalog/MODIS_061_MOD13A2}{MOD13A2};
(iii) Relative Humidity and (iv) Wind Speed from 
\href{https://developers.google.com/earth-engine/datasets/catalog/ECMWF_ERA5_LAND_HOURLY}{ERA5-Land} fields; 
(v) Land Use/Land Cover (LULC) from 
\href{https://developers.google.com/earth-engine/datasets/catalog/GOOGLE_DYNAMICWORLD_V1}{Dynamic World V1}; 
(vi) Soil Moisture from 
\href{https://developers.google.com/earth-engine/datasets/catalog/ISRIC_SoilGrids250m_v2_0}{ISRIC SoilGrids}; 
and (vii) Elevation from the 
\href{https://developers.google.com/earth-engine/datasets/catalog/USGS_SRTMGL1_003}{NASA SRTM} satellite. 
All inputs and CHIRPS-based precipitation targets were accessed and processed using GEE, resampled to a common $1~\mathrm{km} \times 1~\mathrm{km}$ grid (approx.\ $0.0083^\circ$), and referenced to WGS84 (EPSG:4326). This harmonized dataset enables fine-grained, multimodal modeling of monsoon rainfall. Figures~\ref{fig:Karnataka_views} and~\ref{fig:Bihar_views} illustrate input modalities (Sub Figs. (a)-(g)) and targets (Sub Fig. (h)) for the states of Karnataka and Bihar.

\subsection{Experimental Setup}
The proposed model uses a patch-based Attention U-Net architecture for grid-level multi-class precipitation classification. Input samples are extracted as non-overlapping patches of size $32 \times 32$, resulting in the input tensor $\mathbf{X}$ of channel dimension $21$ (7 modalities over 3 months) and spatial size $32 \times 32$. The encoder comprises of four levels with feature dimensions $[64, 128, 256, 512]$, mirrored in the decoder with attention gates for spatially selective up sampling. A dropout rate of 0.3 is applied in the decoder. The model is trained using the Adam optimizer with a learning rate of $1\mathrm{e}{-4}$, weight decay $1\mathrm{e}{-5}$, and focal loss ($\gamma = 2.0$, $\alpha = 1.0$) to mitigate class imbalance. Training runs for up to 50 epochs with batch size 16 and early stopping. A spatially stratified 70\%/15\%/15\% split is used for train/val/test sets. All experiments are conducted on a system with NVIDIA RTX 4500 Ada 24GB GPU, Intel i7 processor using PyTorch and CUDA 12.4. Model performance is evaluated using accuracy, macro precision, recall, and F1-score, including class-wise metrics. In Tables ~\ref{tab:ind_vs_multi} and \ref{tab:model_classwise_f1}, the $-$ indicated absence of that specific precipitation class in that state.
\begin{figure*}[t]
  \centering
  \begin{subfigure}[b]{0.23\textwidth}
    \includegraphics[width=\textwidth, trim=20 20 6 18, clip]{PredictCompare_Fig/Karnataka_GT_Precipitation_Sep_2024.png}
    \caption{\scriptsize Ground Truth}
  \end{subfigure}
  \hspace{3pt}
  \begin{subfigure}[b]{0.225\textwidth}
    \includegraphics[width=\textwidth, trim=20 20 6 18, clip]{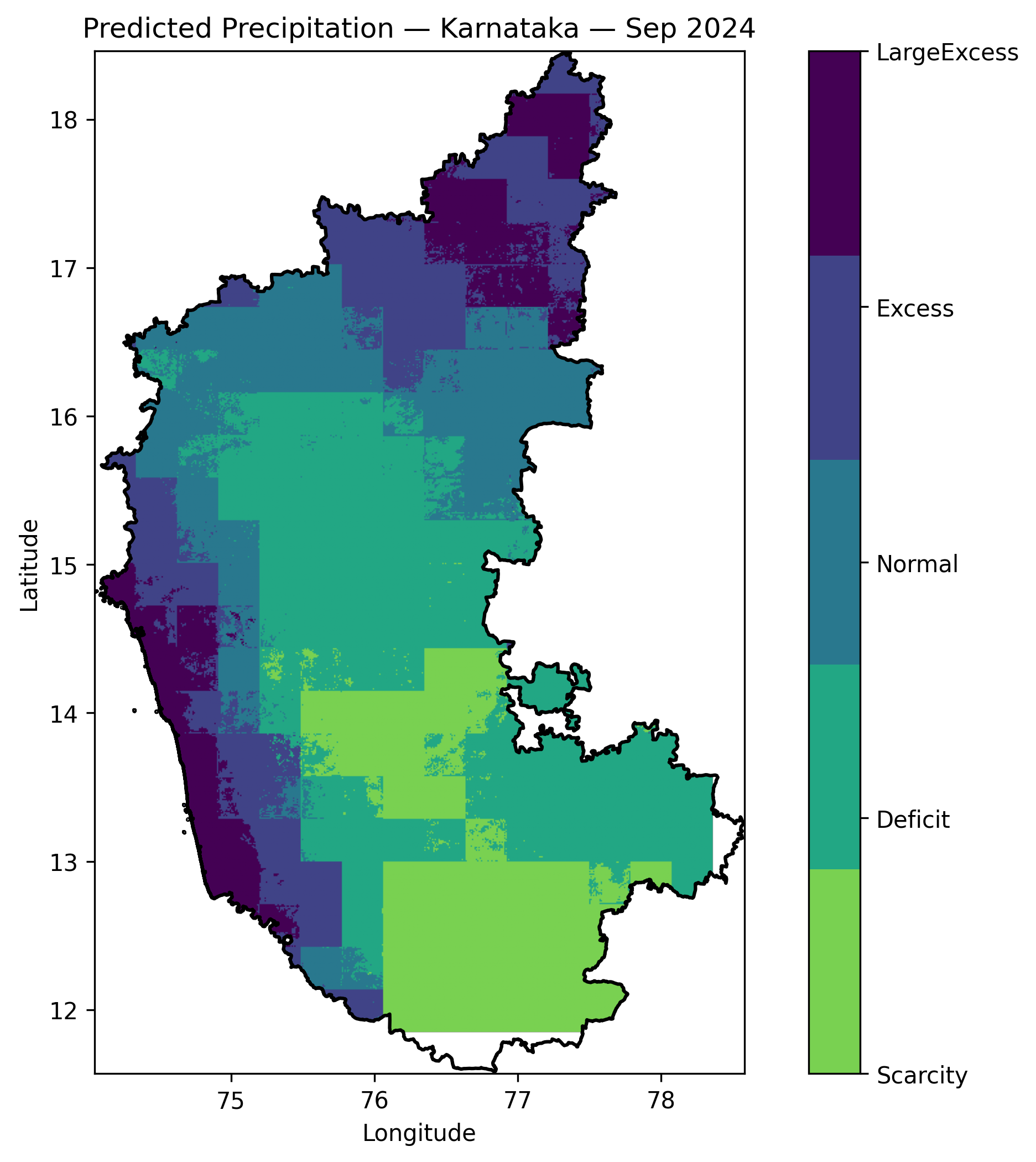}
    \caption{\scriptsize ViT}
  \end{subfigure}
  \hspace{3pt}
  \begin{subfigure}[b]{0.21\textwidth}
    \includegraphics[width=\textwidth, trim=20 20 6 18, clip]{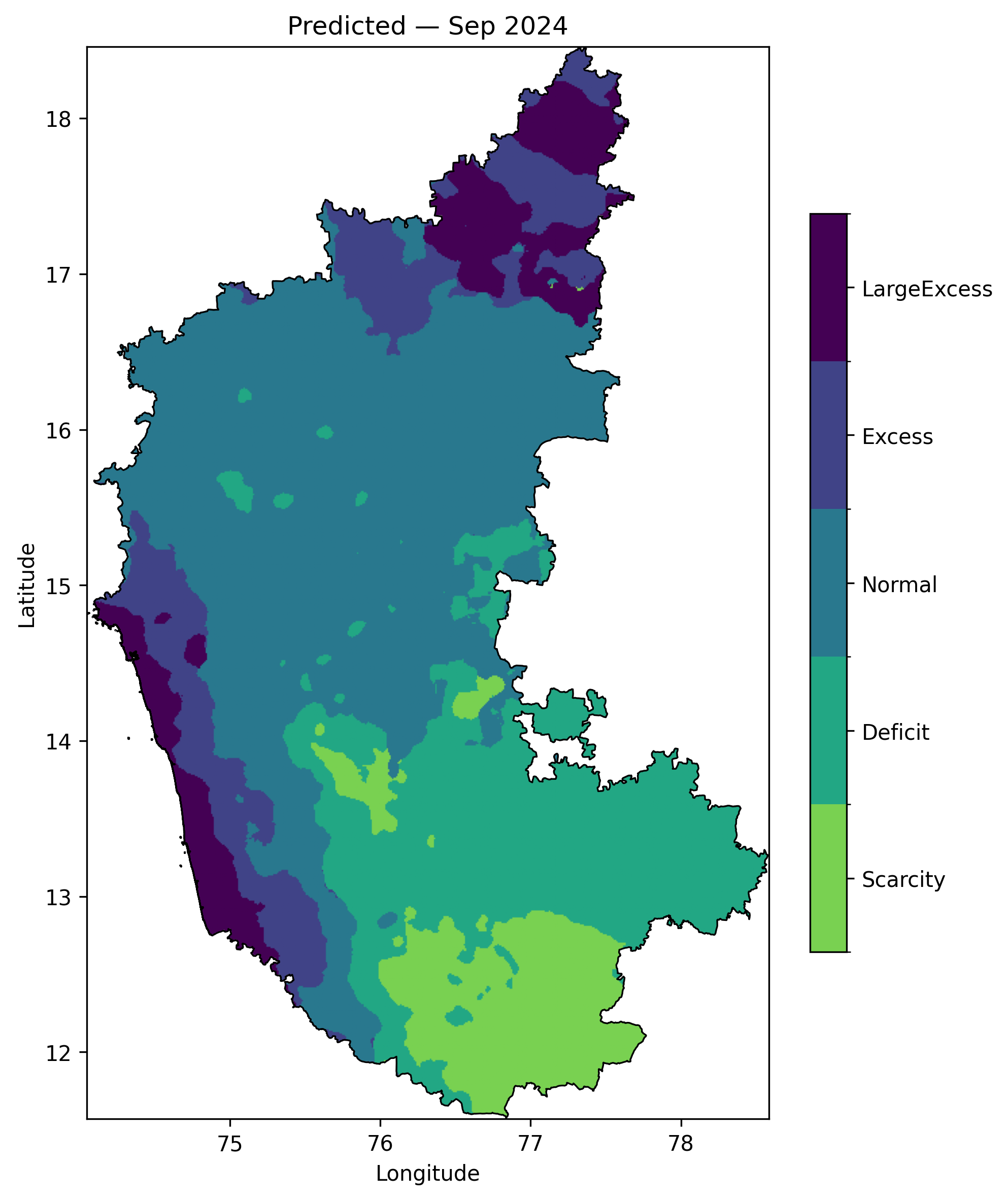}
    \caption{\scriptsize ResNet18}
  \end{subfigure}
  \hspace{3pt}
  \begin{subfigure}[b]{0.225\textwidth}
    \includegraphics[width=\textwidth, trim=20 20 6 18, clip]{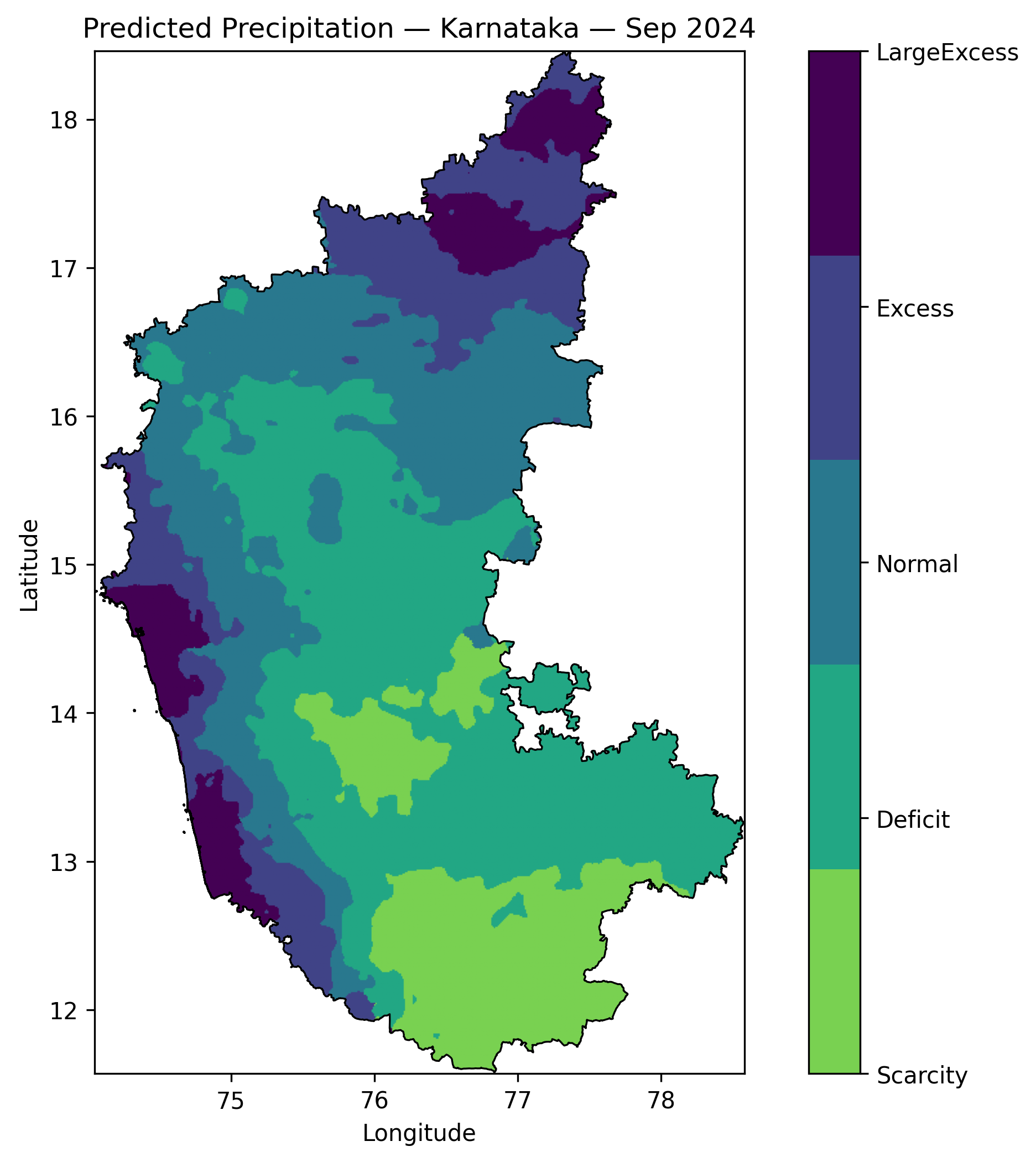}
    \caption{\scriptsize Proposed}
  \end{subfigure}

  \begin{subfigure}[b]{0.23\textwidth}
    \includegraphics[width=\textwidth, trim=20 20 6 18, clip]{PredictCompare_Fig/Bihar_GT_Precipitation_Sep_2024.png}
    \caption{\scriptsize Ground Truth}
  \end{subfigure}
  \hspace{3pt}
  \begin{subfigure}[b]{0.225\textwidth}
    \includegraphics[width=\textwidth, trim=20 20 6 18, clip]{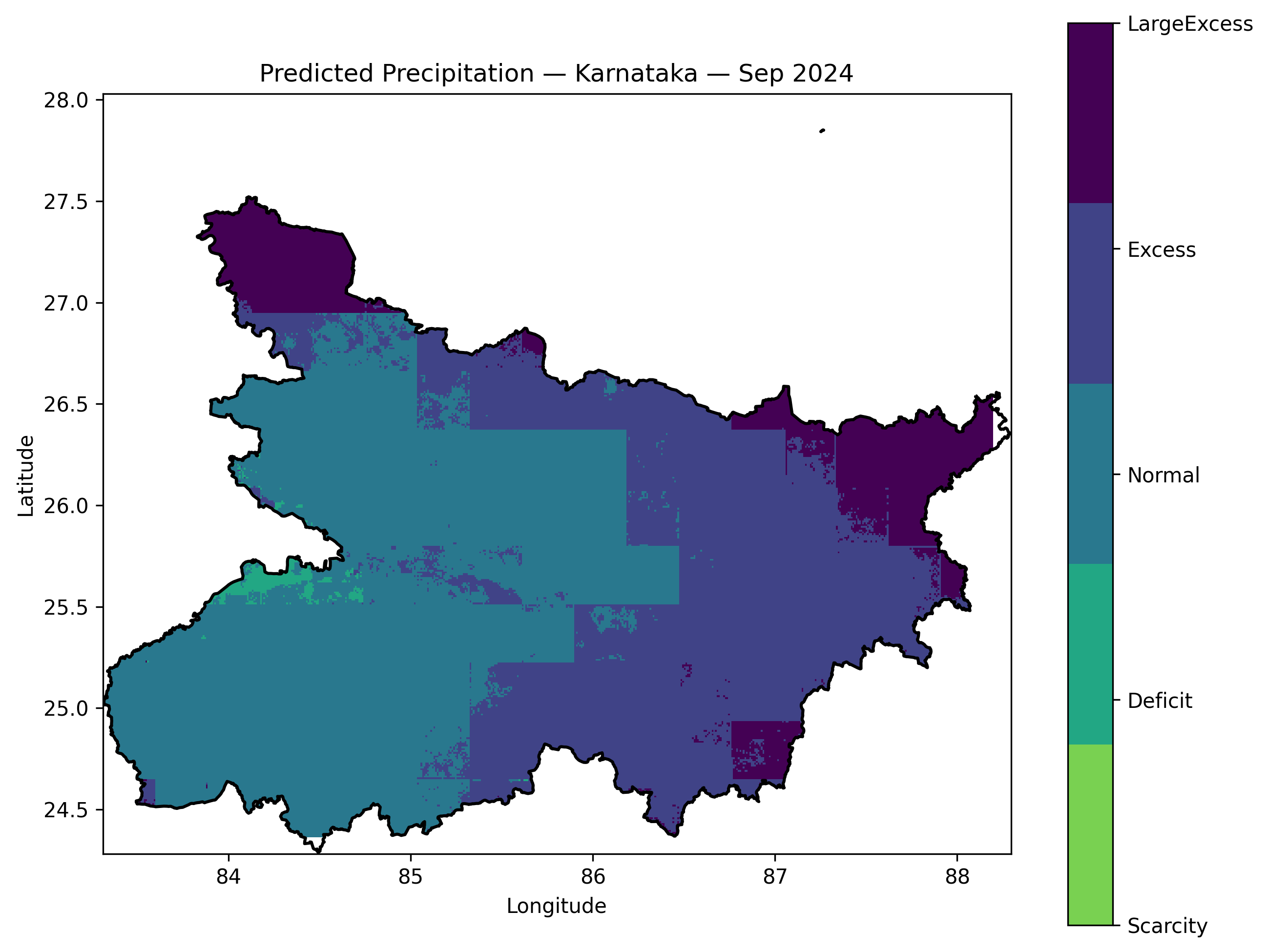}
    \caption{\scriptsize ViT}
  \end{subfigure}
  \hspace{3pt}
  \begin{subfigure}[b]{0.23\textwidth}
    \includegraphics[width=\textwidth, trim=20 20 6 18, clip]{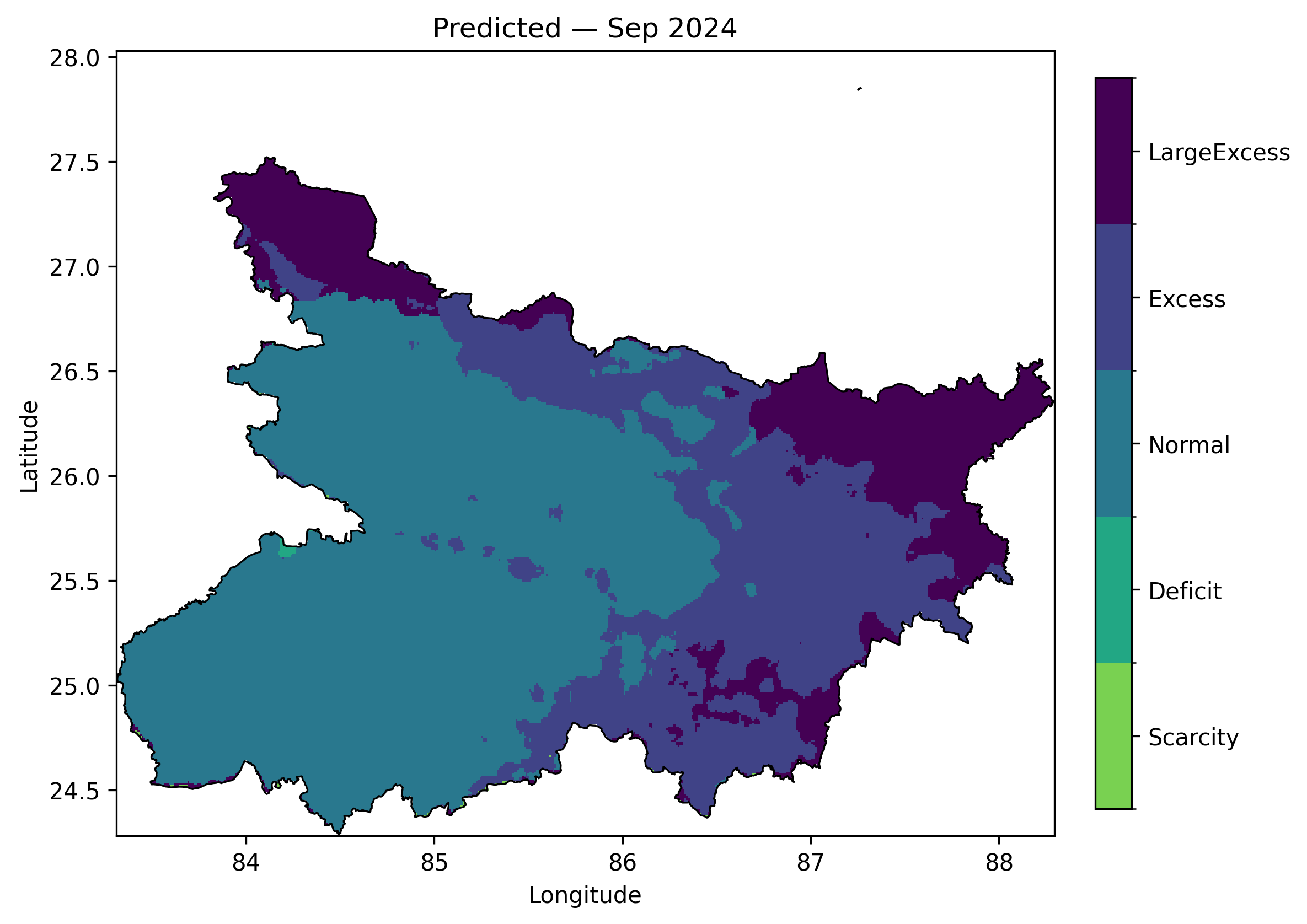}
    \caption{\scriptsize ResNet18}
  \end{subfigure}
  \hspace{3pt}
  \begin{subfigure}[b]{0.225\textwidth}
    \includegraphics[width=\textwidth, trim=20 20 6 18, clip]{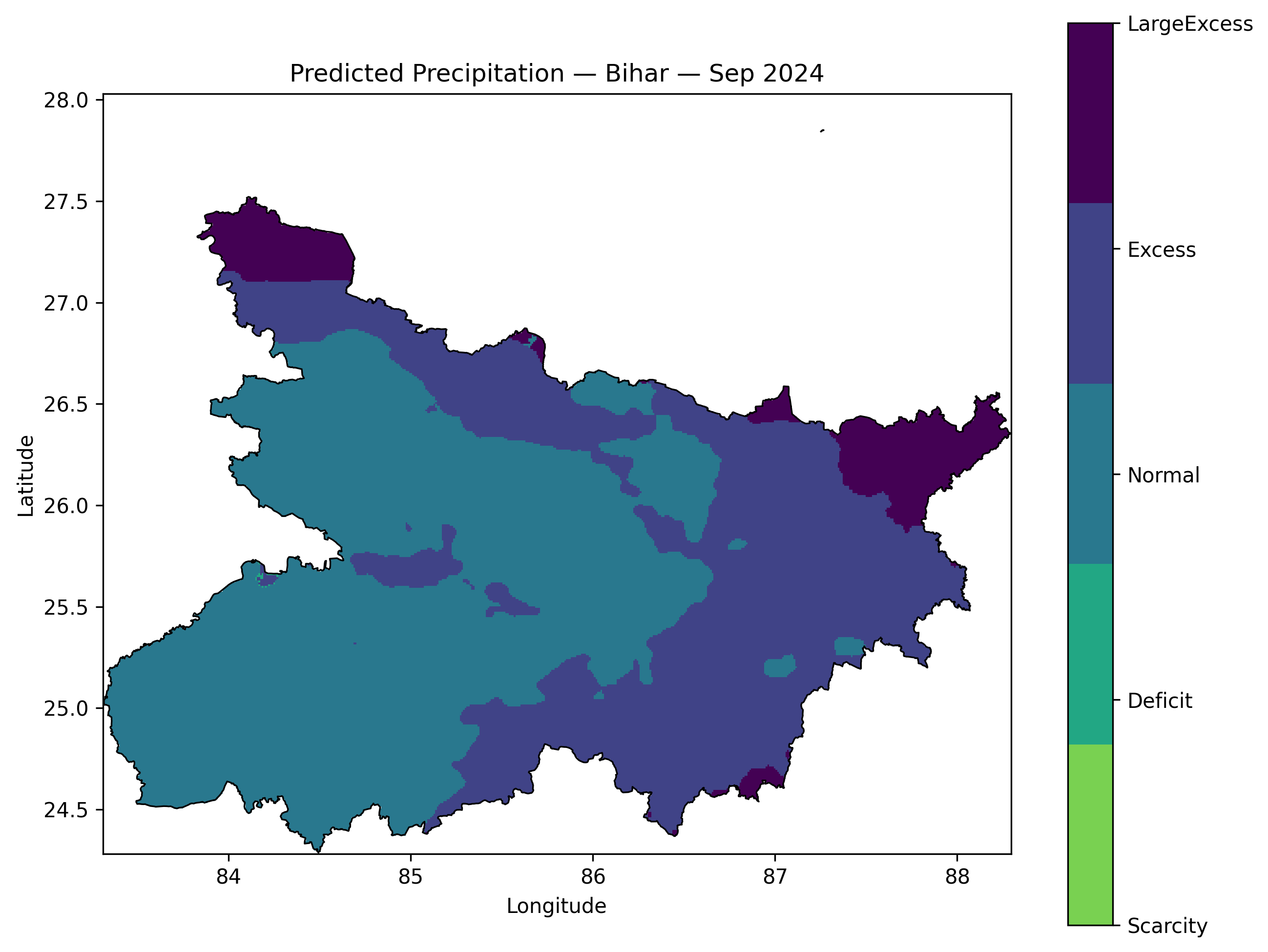}
    \caption{\scriptsize Proposed}
  \end{subfigure}
  \caption{Visualization of Ground truth class labels vs ViT vs ResNet18 vs Proposed Method for Karnataka(a-d) and Bihar(e-h).}
  \label{fig:KR_BHR_Compare}
\vspace{-0.5cm}
\end{figure*}

\begin{table}[t]
\centering
\setlength{\tabcolsep}{4pt}
\caption{Ablation Study of Loss Function Components}
\label{tab:loss_ablation}
\scriptsize
\begin{tabular}{l l c c l c c }
\hline
\textbf{Loss} & \textbf{State} & \textbf{Accuracy} & \textbf{F1 Score} 
              & \textbf{State} & \textbf{Accuracy} & \textbf{F1 Score} \\
\hline
    Focal $\mathcal{L}_{\mathrm{FL}}$        
    & \multirow{3}{*}{Bihar}     & 0.9161   & 0.9144  
    & \multirow{3}{*}{Karnataka} & 0.8700   & 0.8800\\
    Dice $\mathcal{L}_{\mathrm{Dice}}$         
    & & 0.9243   & 0.9218 
    & & 0.9000 & 0.8900\\
    \textbf{Proposed} ($\mathcal{L}_{\mathrm{FL}} + \mathcal{L}_{\mathrm{Dice}}$)   
    & & \textbf{0.9322}  & \textbf{0.9302}  
    & & \textbf{0.9027} & \textbf{0.9036}\\
    
 \hline
     Focal $\mathcal{L}_{\mathrm{FL}}$        
    & \multirow{3}{*}{Assam}     & \textbf{0.9183}   & \textbf{0.9179} 
    & \multirow{3}{*}{Kerala} & 0.7386   & 0.7345 \\
    Dice $\mathcal{L}_{\mathrm{Dice}}$         
    & & 0.9183  & 0.9179 
    & & 0.7521 & 0.7516 \\
    \textbf{Proposed} ($\mathcal{L}_{\mathrm{FL}} + \mathcal{L}_{\mathrm{Dice}}$)   
    & & 0.9045  & 0.9038  
    & & \textbf{0.7764} & \textbf{0.7649}\\
\hline
\end{tabular}
\end{table}
\begin{figure*}[h]
    \centering

    \includegraphics[width=0.32\textwidth]{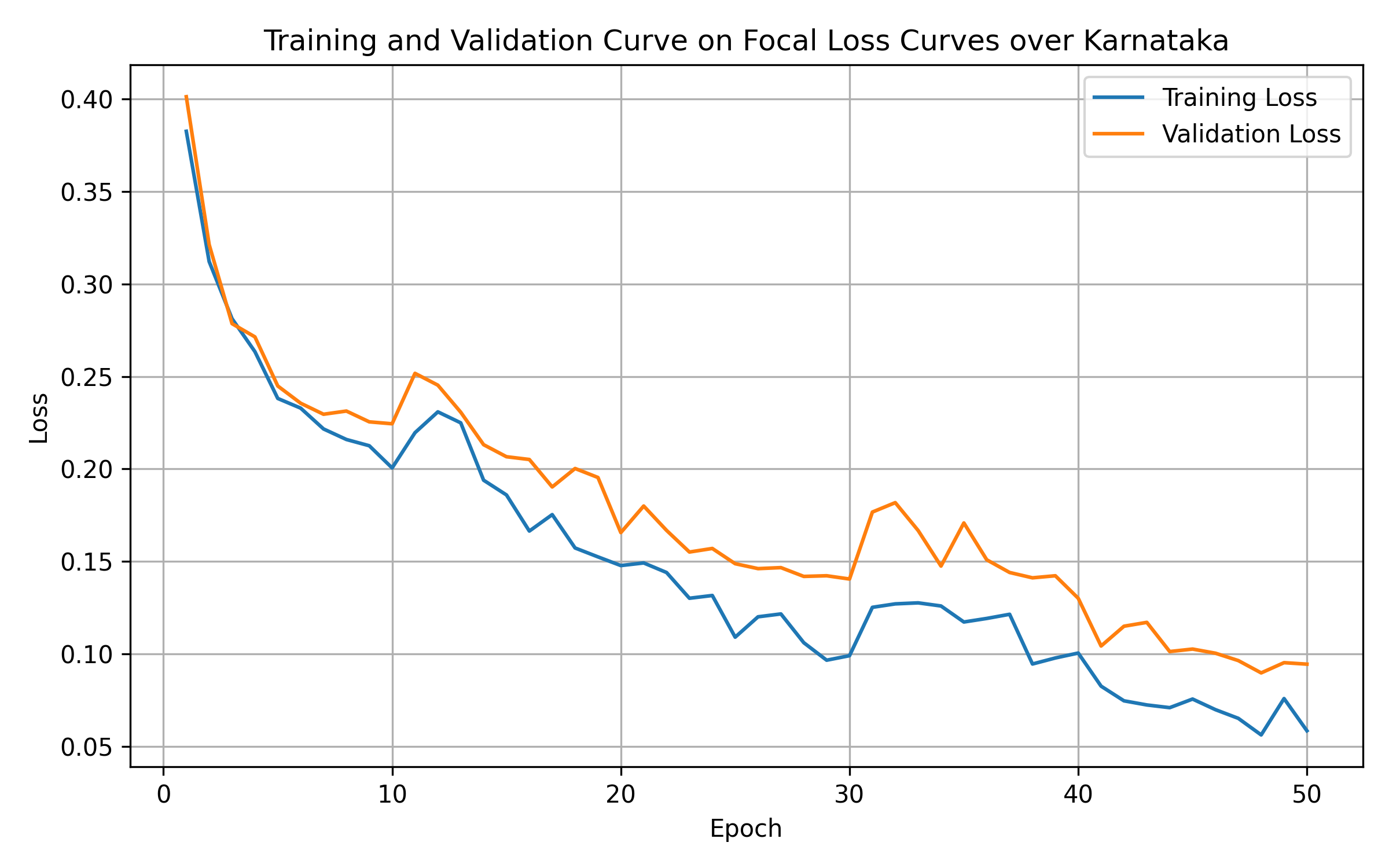}
    \includegraphics[width=0.32\textwidth]{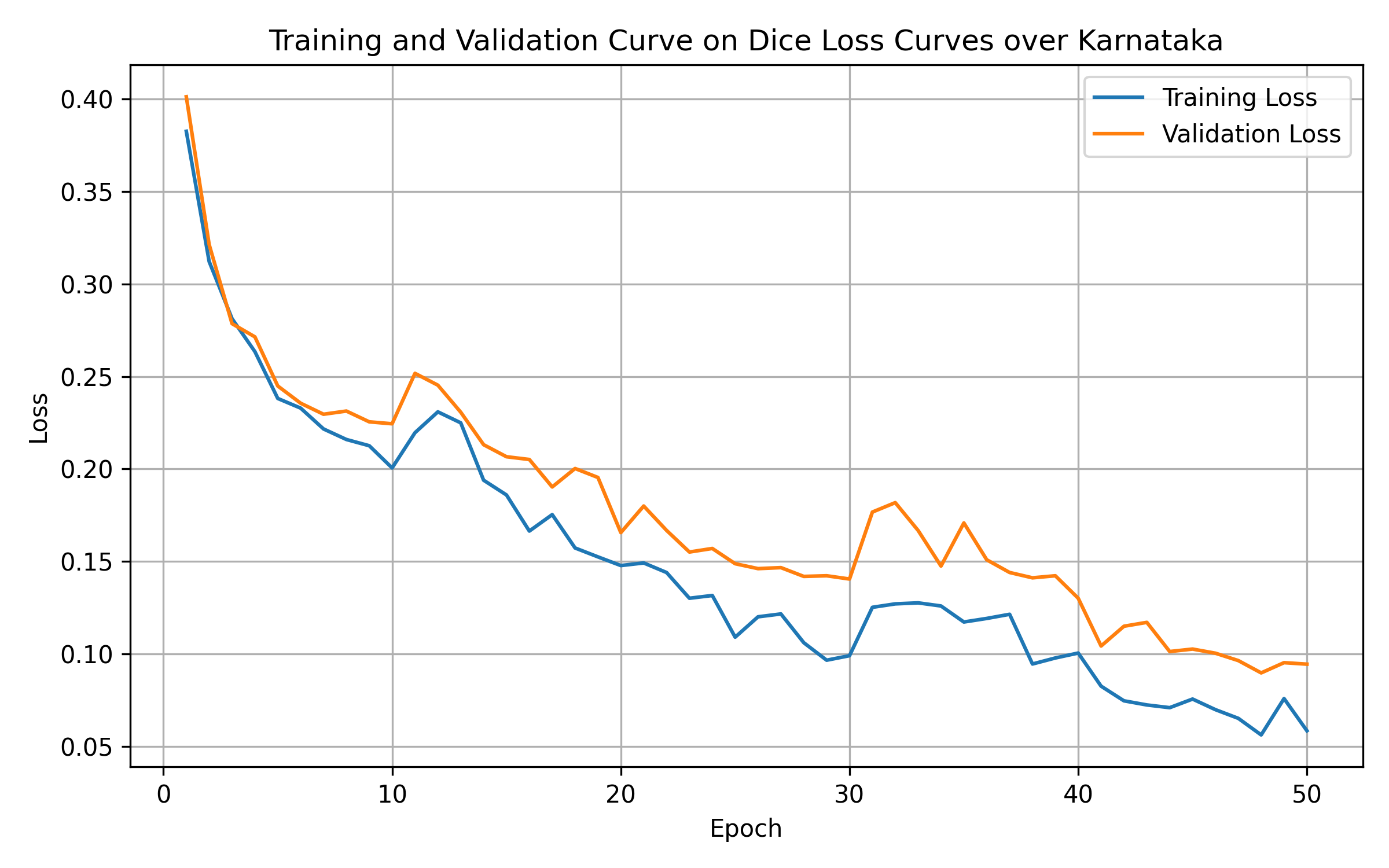}
    \includegraphics[width=0.32\textwidth]{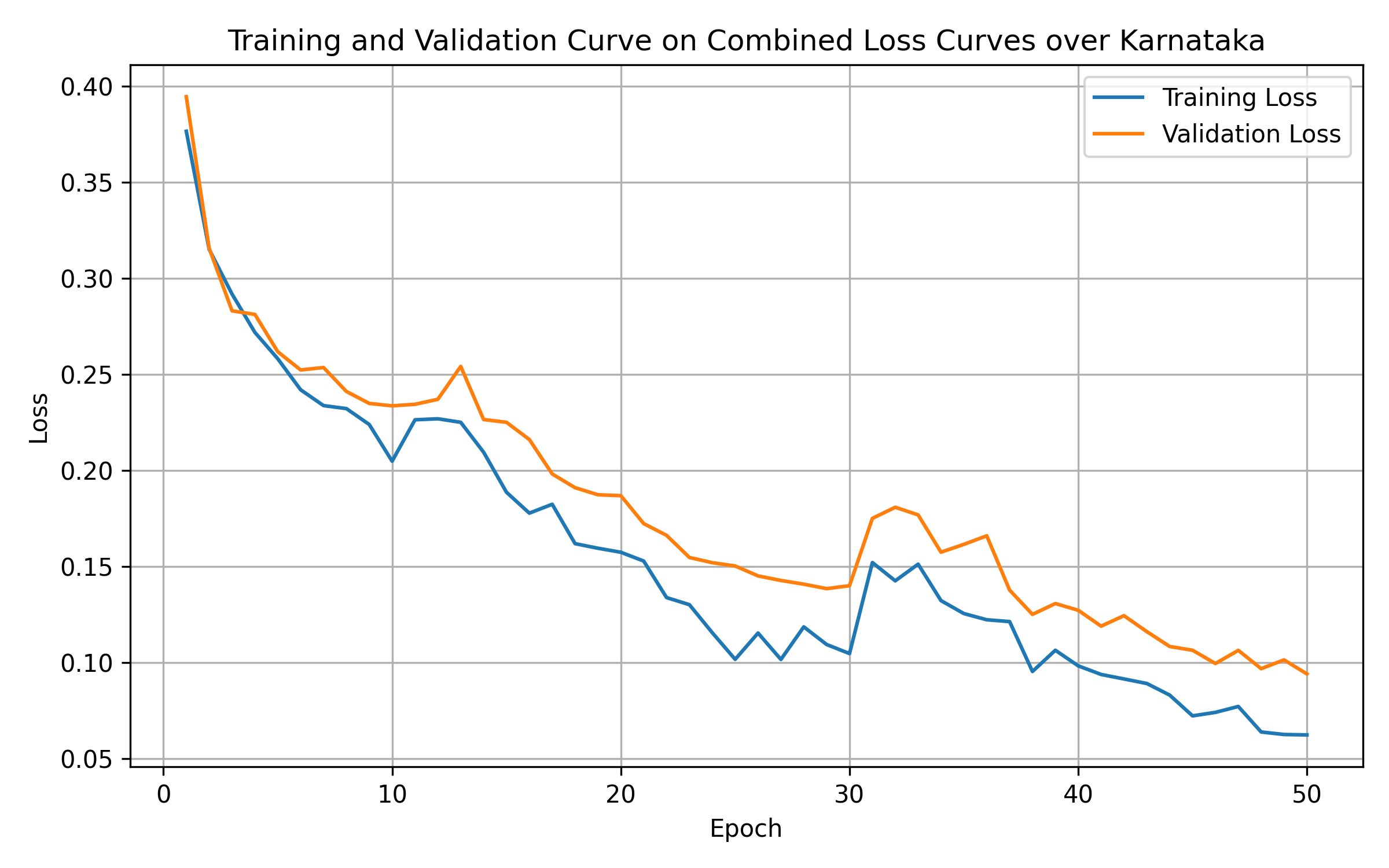} \\

    \includegraphics[width=0.32\textwidth]{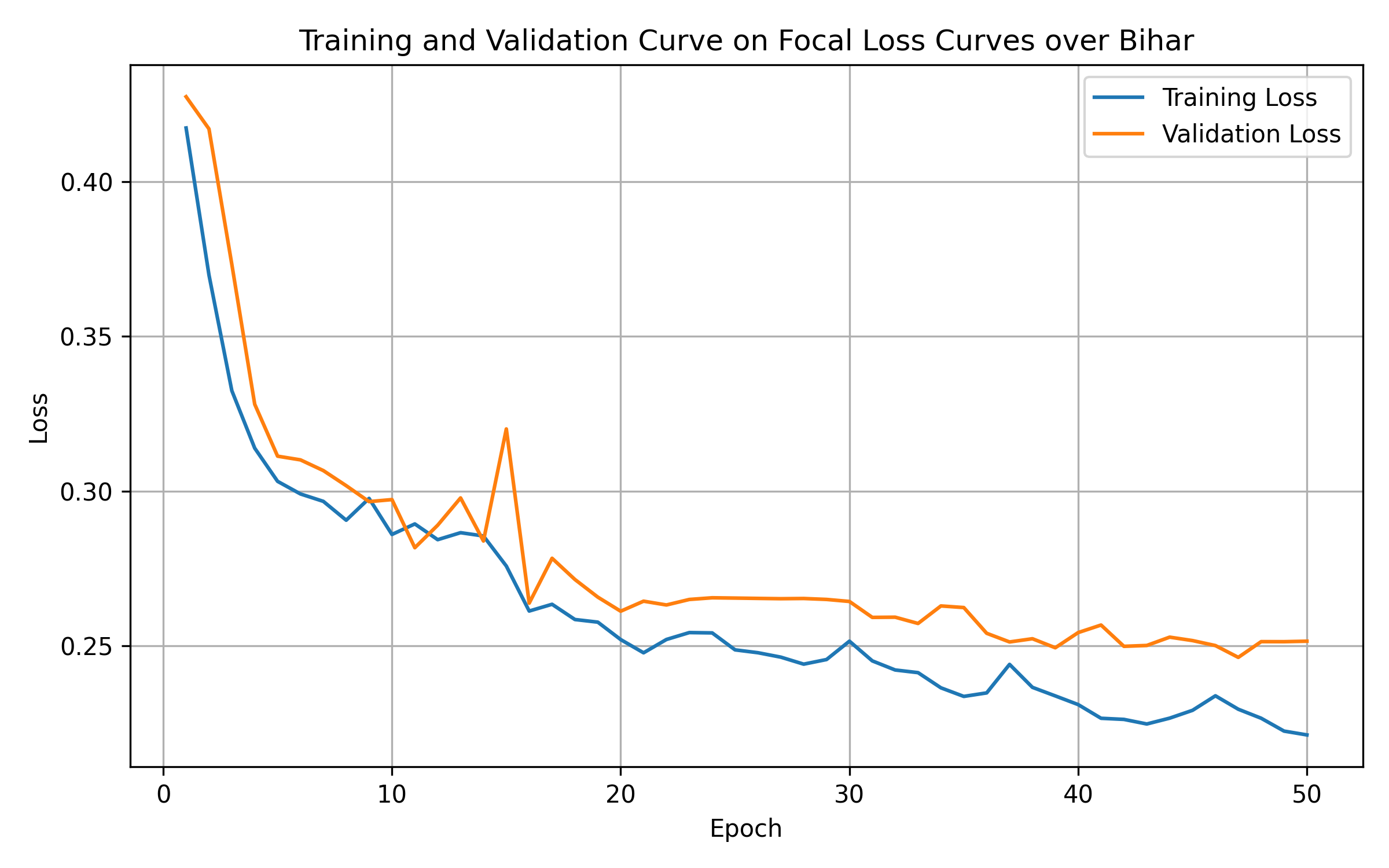}
    \includegraphics[width=0.32\textwidth]{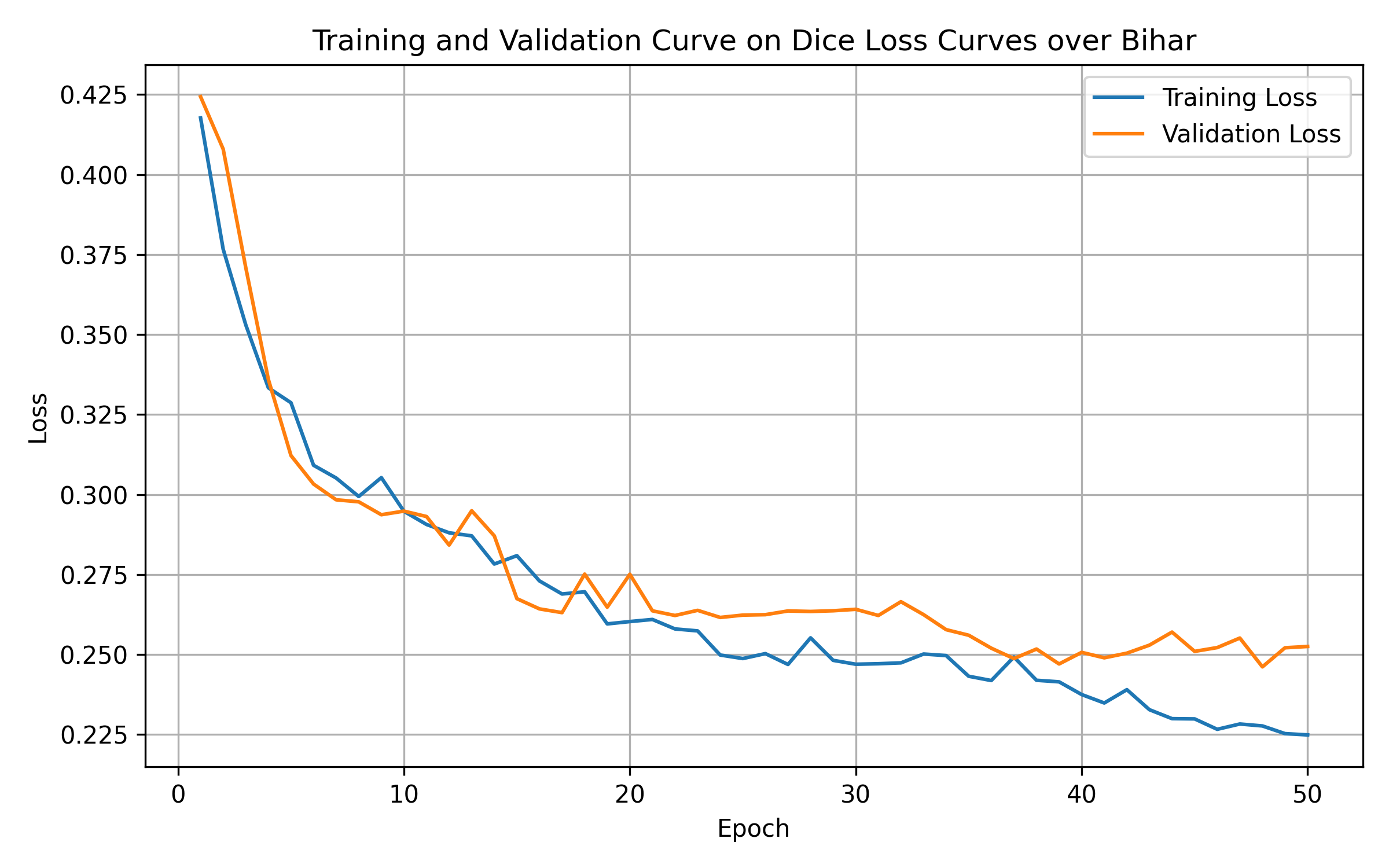}
    \includegraphics[width=0.32\textwidth]{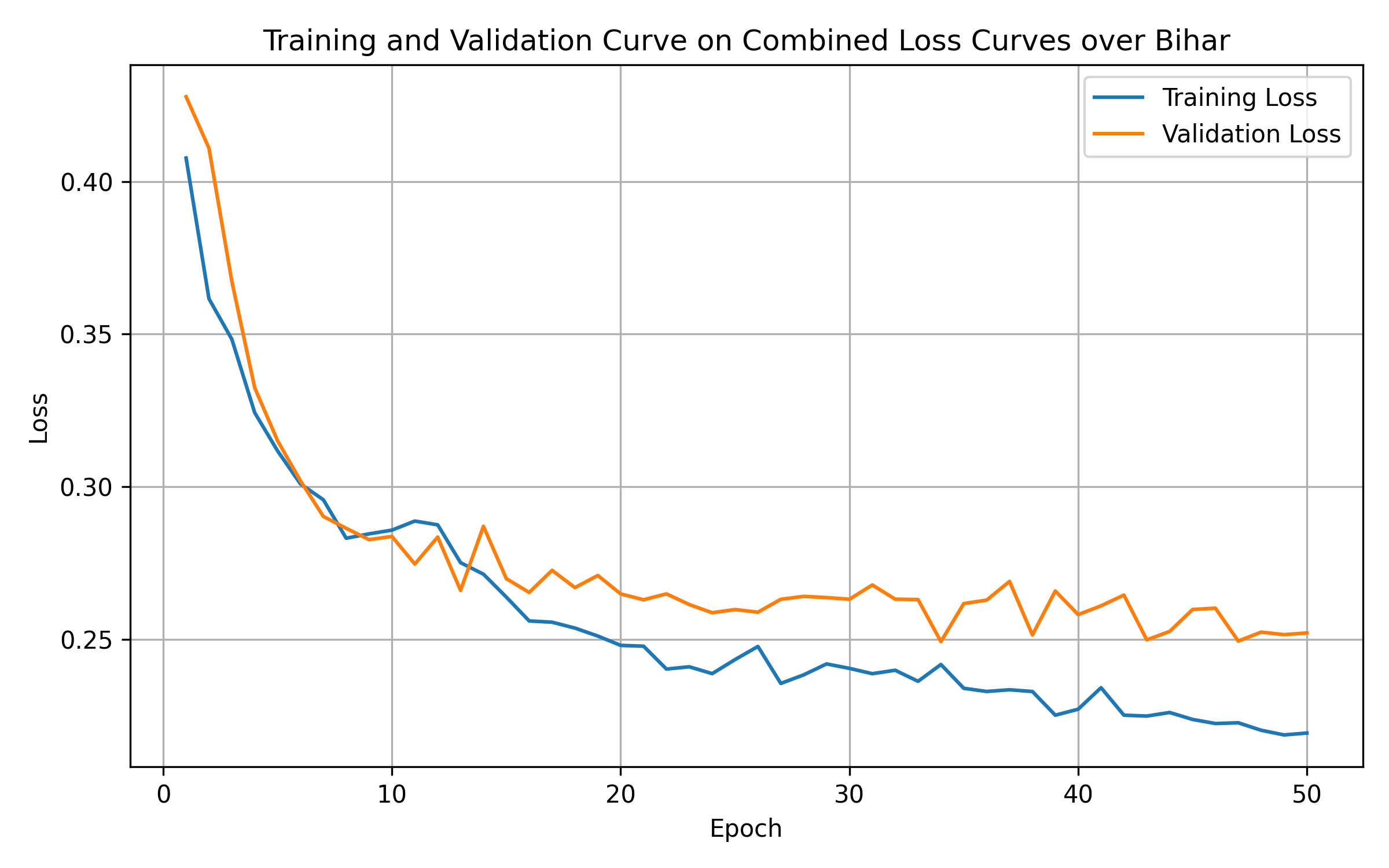} \\

    \makebox[0.32\textwidth]{(a) Focal Loss}
    \makebox[0.32\textwidth]{(b) Dice Loss}
    \makebox[0.32\textwidth]{(c) Combined Loss}

    \caption{Training and validation loss curves for states of Karnataka (top row) and Bihar (bottom row) corresponding to (a) Focal (b) Dice and (c) Combined Losses}
    \label{fig:loss_curves_all_states}
\end{figure*}

\subsection{Ablation Study}
\textit{Significance of Multimodality Integration}: To assess the contribution of each modality, we conducted an ablation study where input modalities were evaluated individually using the proposed Attention U-Net architecture. Table~\ref{tab:ind_vs_multi} compares single-modality performance with that of the multimodal model across five Indian states. We observed that certain modalities individually contribute more significantly to precipitation prediction.  Table~\ref{tab:ind_vs_multi} shows that humidity, wind speed, and LST emerged as the most informative features consistently yielding higher accuracies and F1 scores relative to other inputs, with humidity and wind speed performing well in Bihar and Himachal Pradesh, and humidity and LST in Kerala.  However, these unimodal models struggled to generalize across all rainfall classes and regions. In contrast, the multimodal model consistently outperformed unimodal variants in both accuracy and weighted F1, especially for extreme categories such as \textit{Scarcity} in Himachal Pradesh and Karnataka, and \textit{Large Excess} in coastal Karnataka and northern Bihar (Figure~\ref{fig:KR_BHR_Compare}(d,h)). These results underscore the value of integrating diverse geospatial signals such as thermal, vegetative, atmospheric, and topographic measurements within a unified spatio-temporal framework.

\begin{table*}[t]
\centering
\scriptsize
\caption{Comparative Performance of Existing and Proposed Approach}
\label{tab:model_classwise_f1}
\renewcommand{\arraystretch}{1.0}
\setlength{\tabcolsep}{1pt}
\begin{threeparttable}
\begin{tabular}{@{}llccccccc@{}}
\toprule
\multirow{2}{*}{\textbf{State}} & \multirow{2}{*}{\textbf{Model}} & 
  \multirow{2}{*}{\makecell{\textbf{Overall}\\\textbf{Accuracy}}} & 
  \multirow{2}{*}{\makecell{\textbf{Weighted}\\\textbf{F1 Score}}} & 
  \multicolumn{5}{c}{\textbf{Class-wise F1 Score}} \\
\cmidrule(lr){5-9}
  &  &  &  & \textbf{Scarcity} & \textbf{Deficit} & \textbf{Normal} & \textbf{Excess} & \textbf{Large Excess} \\
\midrule

\multirow{8}{*}{Assam} 
  & SVM        & 0.6764 & 0.6589 & 0.6432 & 0.7481 & 0.4275 & 0.1277 & -- \\
  & RF         & 0.8323 & 0.8328 & 0.8496 & 0.8416 & 0.7667 & 0.5103 & -- \\
  & XGBoost    & 0.8680 & 0.8707 & 0.8916 & 0.8866 & 0.7769 & 0.2996 & -- \\
  & 1DCNN      & 0.6521 & 0.6581 & 0.7022 & 0.6790 & 0.5672 & 0.4545 & -- \\
  & ViT        & 0.7495 & 0.7472 & 0.7933 & 0.8149 & 0.4189 & 0.0836 & -- \\
  & ResNet18   & 0.7535 & 0.7573 & 0.7350 & 0.7846 & 0.7441 & 0.2281 & -- \\
  & \textbf{\textbf{Proposed} } 
              & \textbf{0.9045} & \textbf{0.9038} 
              & \textbf{0.9245} & \textbf{0.8932} 
              & \textbf{0.8996} & \textbf{1.0000} & -- \\
\midrule

\multirow{8}{*}{Bihar} 
  & SVM        & 0.8573 & 0.8534 & --      & 0.0000 & 0.9141 & 0.8275 & 0.6325 \\
  & RF         & 0.9112 & 0.9107 & --      & \textbf{0.8525} & \textbf{0.9514} & 0.8931 & 0.7130 \\
  & XGBoost  
              & \textbf{0.9419} & \textbf{0.9406} & -- 
              & 0.4361 & 0.9493 & \textbf{0.9379} & -- \\
  & 1DCNN      & 0.8373 & 0.8403 & --      & 0.2078 & 0.8747 & 0.7963 & 0.8234 \\
  & ViT        & 0.7670 & 0.7710 & --      & 0.3579 & 0.8452 & 0.7076 & 0.6111 \\
  & ResNet18   & 0.7232 & 0.7347 & --      & 0.0000 & 0.8081 & 0.6993 & 0.4607 \\
  & \textbf{Proposed}  & 0.9322 & 0.9302 & --      & 0.0000 & 0.9153 & 0.9335 & \textbf{0.9758} \\
\midrule

\multirow{8}{*}{Himachal} 
  & SVM        & 0.7249 & 0.7240 & 0.9058 & 0.1503 & 0.3714 & 0.5649 & 0.8461 \\
  & RF         & 0.6927 & 0.6980 & 0.9434 & 0.4680 & 0.4197 & 0.4553 & 0.8198 \\
  & XGBoost    & 0.7190 & 0.7150 & 0.8772 & 0.4421 & 0.4723 & 0.4769 & 0.8474 \\
  & 1DCNN      & 0.6451 & 0.6473 & 0.7211 & 0.6193 & 0.5703 & 0.5373 & 0.7595 \\
  & ViT
              & 0.5622 & 0.5450 & \textbf{0.9735} & 0.0000 
              & 0.4315 & 0.1466 & 0.6948 \\
  & ResNet18   & 0.5790 & 0.5813 & 0.9665 & 0.0000 & 0.5452 & 0.2655 & 0.6899 \\
  & \textbf{Proposed}
              & \textbf{0.8802} & \textbf{0.8837} 
              & 0.9612 & \textbf{0.9681} 
              & \textbf{0.8690} & \textbf{0.6510}
              & \textbf{0.8506} \\
\midrule

\multirow{8}{*}{Karnataka} 
  & SVM        & 0.6812 & 0.6552 & 0.8713 & 0.8497 & 0.4585 & 0.5705 & 0.1114 \\
  & RF         & 0.7961 & 0.7990 & 0.9105 & 0.8956 & 0.6464 & 0.6670 & 0.7165 \\
  & XGBoost    & 0.8574 & 0.8590 & 0.9228 & \textbf{0.9122} 
              & 0.7686 & 0.7972 & 0.8060 \\
  & 1DCNN      & 0.7111 & 0.7128 & 0.7200 & 0.7729 & 0.6664 & 0.6396 & 0.6725 \\
  & ViT        & 0.7536 & 0.7555 & 0.9305 & \textbf{0.9122} 
              & 0.5466 & 0.4415 & 0.7002 \\
  & ResNet18   & 0.5961 & 0.6018 & 0.7042 & 0.7736 & 0.4561 & 0.4452 & 0.2730 \\
  & \textbf{Proposed}
              & \textbf{0.9027} & \textbf{0.9036}
              & \textbf{0.9804} & 0.8219
              & \textbf{0.8619} & \textbf{0.8730}
              & \textbf{0.9363} \\
\midrule

\multirow{8}{*}{Kerala} 
  & SVM        & 0.7857 & 0.7946 & 0.8533 & 0.6769 & 0.6631 & --      & -- \\
  & RF         & 0.7701 & 0.7827 & 0.8254 & 0.6575 & 0.8204 & --      & -- \\
  & XGBoost    & \textbf{0.7865} & \textbf{0.7985}
              & 0.8345 & 0.6798 & \textbf{0.8748} & -- & -- \\
  & 1DCNN      & 0.7151 & 0.7204 & 0.7368 & 0.7227 & 0.5729 & --      & -- \\
  & ViT        & 0.7234 & 0.7873 & 0.8374 & 0.7822 & 0.3528 & --      & -- \\
  & ResNet18   & 0.6105 & 0.6398 & 0.6641 & 0.6708 & 0.3157 & --      & -- \\
  & \textbf{Proposed}  & 0.7764 & 0.7649 & 0.6561 & \textbf{0.7869}
              & 0.9527 & --      & -- \\
\bottomrule
\end{tabular}
\begin{tablenotes}
\item $-$ indicates absence of that specific precipitation class in a state.
\end{tablenotes}
\end{threeparttable}
\vspace{-0.5cm}
\end{table*}

\textit{Loss Function Ablation}: Table~\ref{tab:loss_ablation} presents the impact of different loss functions across four states. The combined loss formulation ($\mathcal{L}{\mathrm{FL}} + \mathcal{L}{\mathrm{Dice}}$) consistently achieves the highest accuracy and weighted F1 in Bihar, Karnataka, and Kerala, while maintaining competitive performance in Assam. This highlights the complementary benefits of focal loss for addressing class imbalance and dice loss for enhancing spatial consistency. Figure~\ref{fig:loss_curves_all_states} visualizes training and validation loss curves for each loss function variant. Models trained with combined loss exhibited faster convergence and lower final losses compared to focal-only and dice-only configurations. These results underscore the importance of composite loss design in improving classification performance across diverse regional rainfall patterns.

\subsection{Comparative Performance Analysis}
Table~\ref{tab:model_classwise_f1} benchmarks the proposed model against classical machine learning models like support vector machines (SVM), random forest (RF), and XGBoost, shallow neural networks like 1-dimensional convolution neural networks (1D CNN), and deep architectures like ResNet18 \cite{ResNet}, vision transformers (ViT) \cite{ViT}. The proposed method consistently outperforms all baselines across five states in both overall accuracy and weighted F1 score, highlighting its robustness to regional rainfall heterogeneity. While XGBoost performs competitively in homogeneous regions of Bihar and Kerala, its performance degrades in topographically diverse states such as Himachal Pradesh and Karnataka. Notably, the proposed method yields substantial improvements in rare classes such as \textit{Scarcity} and \textit{Large Excess}, attributable to its use of focal loss. Furthermore, it captures sharp spatial transitions more effectively than vision transformers (Figure~\ref{fig:KR_BHR_Compare}(h) vs.~\ref{fig:KR_BHR_Compare}(f)). These results underscore the value of attention-based spatio-temporal fusion of multi-source geophysical inputs for high-resolution localized precipitation forecasting.

\section{Conclusion}
This work introduces a multimodal deep learning framework for grid-level precipitation classification by leveraging heterogeneous satellite-derived geospatial inputs. Through an attention-enhanced U-Net architecture equipped with joint optimization of focal and Dice loss, the model effectively fuses multimodal Earth observation data across time, capturing spatio-temporal patterns that traditional and unimodal models fail to exploit. Empirical results across five climatically diverse Indian states show that the proposed approach consistently outperforms both individual modality baselines and other existing models such as XGBoost, ViT, and ResNet18, particularly in the underrepresented extreme rainfall categories. Beyond precipitation classification, the framework holds promise as a regional foundation model for Earth observation prediction. Its modular design enables adaptation to related forecasting tasks such as flood or drought risk mapping and scalable deployment across geographies. Future work will explore its extension to nowcasting settings and continuous precipitation estimation, with an emphasis on pan-India and cross-national datasets.


\begin{thebibliography}{10}
\providecommand{\url}[1]{\texttt{#1}}
\providecommand{\urlprefix}{URL }
\providecommand{\doi}[1]{https://doi.org/#1}

\bibitem{ADERYANI2022128463}
Short-term rainfall forecasting using machine learning-based approaches of pso-svr, lstm and cnn. Journal of Hydrology  \textbf{614},  128463 (2022)

\bibitem{nowcasting_cao}
Cao, Y., Chen, L., Wu, J., Feng, J.: Enhancing nowcasting with multi-resolution inputs using deep learning: Exploring model decision mechanisms. Geophysical Research Letters  \textbf{52}(4),  e2024GL113699 (2025)

\bibitem{Dash2024SpatialPrecip}
Dash, S., et~al.: Changes in the spatial variability of extreme precipitation characteristics across peninsular india. Discover Geoscience  \textbf{2}(1), ~17 (Jun 2024)

\bibitem{ViT}
Dosovitskiy, A., et~al.: An image is worth 16x16 words: Transformers for image recognition at scale. In: 9th International Conference on Learning Representations, {ICLR} 2021, Virtual Event, Austria, May 3-7, 2021 (2021)

\bibitem{Maity2013HydrologyMonsoon}
Falga, R., Wang, C.: The rise of indian summer monsoon precipitation extremes and its correlation with long-term changes of climate and anthropogenic factors. Scientific Reports  \textbf{12}(1),  11985 (Jul 2022)

\bibitem{Ghosh2011RainfallExtremes}
Ghosh, S., Das, D., Kao, S.C., Ganguly, A.R.: Lack of uniform trends but increasing spatial variability in observed indian rainfall extremes. Nature Climate Change  \textbf{2}(2),  86--91 (Feb 2012)

\bibitem{ResNet}
He, K., Zhang, X., Ren, S., Sun, J.: Deep residual learning for image recognition. In: 2016 IEEE Conference on Computer Vision and Pattern Recognition (CVPR). pp. 770--778 (2016)

\bibitem{Hou2014GPM}
Hou, A.Y., et~al.: The global precipitation measurement mission. Bulletin of the American Meteorological Society  \textbf{95}(5),  701--722 (2014)

\bibitem{gmd-17-709-2024}
Jain, D., et~al.: Monsoon mission coupled forecast system version 2.0: model description and indian monsoon simulations. Geoscientific Model Development  \textbf{17}(2),  709--729 (2024)

\bibitem{lam2022learning}
Lam, R., et~al.: Learning skillful medium-range global weather forecasting. Science  \textbf{377}(6611),  1111--1117 (2022)

\bibitem{10.2166/hydro.2024.014}
Latif, S.D., Mohammed, D.O., Jaafar, A.: Developing an innovative machine learning model for rainfall prediction in a semi-arid region. Journal of Hydroinformatics  \textbf{26}(4),  904--914 (03 2024)

\bibitem{inbook}
Li, C., Feng, Y., Sun, T., Zhang, X.: Long term indian ocean dipole (iod) index prediction used deep learning by convlstm. Remote Sensing  \textbf{14}(3) (2022)

\bibitem{focal_loss}
Lin, T.Y., Goyal, P., Girshick, R., He, K., Dollar, P.: Focal loss for dense object detection. In: 2017 IEEE International Conference on Computer Vision (ICCV). pp. 2999--3007 (Oct 2017)

\bibitem{imd2024monsoon}
Mohapatra, M., Jenamani, R., Prakash, S.: Monsoon 2024: A report (2024)

\bibitem{atten_unet}
Oktay, O., et~al.: Attention {U-N}et: Learning where to look for the pancreas (2018)

\bibitem{NEIndiaExtremes2023}
Paul, A.R., Maity, R.: Future projection of climate extremes across contiguous northeast india and bangladesh. Scientific Reports  \textbf{13}(1),  15616 (Sep 2023)

\bibitem{skilful}
Ravuri, S., et~al.: Skilful precipitation nowcasting using deep generative models of radar. Nature  \textbf{597}(7878),  672--677 (Sep 2021)

\bibitem{unet}
Ronneberger, O., Fischer, P., Brox, T.: U-net: Convolutional networks for biomedical image segmentation. In: Medical Image Computing and Computer-Assisted Intervention -- MICCAI 2015. pp. 234--241 (2015)

\bibitem{Roxy2017CentralIndia}
Roxy, M.K., et~al.: A threefold rise in widespread extreme rain events over central india. Nature Communications  \textbf{8}(1), ~708 (Oct 2017)

\bibitem{Tirumani2021Intercomp}
Saikrishna, T.S., Ramu, D.A., Osuri, K.K.: Inter-comparison of high-resolution satellite precipitation products over india during the summer monsoon season. Meteorology and Atmospheric Physics  \textbf{133}(6),  1675--1690 (Dec 2021)

\bibitem{shi2017deep}
Shi, X., et~al.: Deep learning for precipitation nowcasting: A benchmark and a new model. In: Advances in Neural Information Processing Systems. vol.~30 (2017)

\bibitem{dice_loss}
Sudre, C.H., et~al.: Generalised dice overlap as a deep learning loss function for highly unbalanced segmentations. In: Deep Learning in Medical Image Analysis and Multimodal Learning for Clinical Decision Support. pp. 240--248 (2017)

\bibitem{Wani2024}
Wani, O.A., et~al.: Predicting rainfall using machine learning, deep learning, and time series models across an altitudinal gradient in the north-western himalayas. Scientific Reports  \textbf{14}(1),  27876 (Nov 2024)

\bibitem{Xu2025}
Xu, P., et~al.: An artificial intelligence-based limited area model for forecasting of surface meteorological variables. Communications Earth {\&} Environment  \textbf{6}(1), ~372 (2025)

\bibitem{XU20251732}
Xu, Y., Yang, Z., Zhang, F., Chen, X., Zhou, H.: A rainfall prediction model based on era5 and elman neural network. Advances in Space Research  \textbf{75}(2),  1732--1746 (2025)

\bibitem{yu2023multisource}
Yu, D., Feng, W., Lin, K., Li, X., Ye, Y., Luo, C., Du, W.: Integrating multi-source data for long sequence precipitation forecasting. Proceedings of the AAAI Conference on Artificial Intelligence  \textbf{39}(27),  28539--28547 (Apr 2025)

\bibitem{ebiak1987model}
Zebiak, S.E., Cane, M.A.: A model el niñoâ€“southern oscillation. Monthly Weather Review  \textbf{115}(10),  2262 -- 2278 (1987)

\bibitem{rs17071123}
Zhang, Z., Song, Q., Duan, M., Liu, H., Huo, J., Han, C.: Deep learning model for precipitation nowcasting based on residual and attention mechanisms. Remote Sensing  \textbf{17}(7) (2025)

\end{thebibliography}
\end{document}